\documentclass[10pt,twocolumn,letterpaper]{article}

\usepackage[pagenumbers]{cvpr} %

\usepackage{amsmath,amsthm}
\definecolor{axisred}{HTML}{B00012}
\definecolor{axisblue}{HTML}{00449E}

\theoremstyle{plain}
\newtheorem{proposition}{Proposition}
\usepackage{xcolor,tikz}

\usepackage{booktabs}    %
\usepackage{multirow}    %
\usepackage{graphicx}    %
\usepackage{colortbl}    %
\usepackage[T1]{fontenc}
\usepackage{newtxtext}
\definecolor{oursbg}{RGB}{232,242,252} %
\usepackage{adjustbox}
\usepackage{tabularx}

\definecolor{cTS}{HTML}{27368F}\definecolor{cD3}{HTML}{C99A1E}\definecolor{cD2}{HTML}{4F86C0}
\definecolor{cCL}{HTML}{9A62B8}\definecolor{cWR}{HTML}{3F9E63}

\newcommand{\mkDt}{\tikz[baseline=-0.2ex]\fill[cD3] (-0.55ex,-0.25ex) rectangle (0.55ex,0.85ex);}
\newcommand{\mkDs}{\tikz[baseline=-0.2ex]\fill[cD2] (-0.65ex,-0.3ex) -- (0.65ex,-0.3ex) -- (0,0.9ex) -- cycle;}
\newcommand{\mkCL}{\tikz[baseline=-0.2ex]\fill[cCL] (0,-0.4ex) -- (0.65ex,0.3ex) -- (0,1.0ex) -- (-0.65ex,0.3ex) -- cycle;}
\newcommand{\mkWR}{\tikz[baseline=-0.2ex]\fill[cWR] (-0.65ex,0.9ex) -- (0.65ex,0.9ex) -- (0,-0.3ex) -- cycle;}

\definecolor{frzBorder}{HTML}{7FA6D6}  \definecolor{frzFill}{HTML}{E3EDF9}
\definecolor{trnBorder}{HTML}{E39B8C}  \definecolor{trnFill}{HTML}{FBE3DD} 
\definecolor{shrinkTxt}{HTML}{2E7D6B}                                    
\definecolor{recenterTxt}{HTML}{C8664A}                                 

\DeclareRobustCommand{\figchip}[3]{%
  \tikz[baseline=(c.base)]\node[draw=#1,fill=#2,rounded corners=1.5pt,
    line width=0.5pt,inner xsep=1.6pt,inner ysep=0.9pt](c){#3};}

\DeclareRobustCommand{\Tch}{\figchip{frzBorder}{frzFill}{$\mathcal{T}$}}
\DeclareRobustCommand{\Stu}{\figchip{trnBorder}{trnFill}{$\mathcal{S}$}}
\DeclareRobustCommand{\TS}{\figchip{frzBorder}{frzFill}{$\mathcal{T}{-}\mathcal{S}$}}
\DeclareRobustCommand{\shrink}{\textcolor{shrinkTxt}{\textbf{shrinkage}}}
\DeclareRobustCommand{\recenter}{\textcolor{recenterTxt}{\textbf{re-centering}}}

\definecolor{cvprblue}{rgb}{0.21,0.49,0.74}
\usepackage[pagebackref,breaklinks,colorlinks,allcolors=cvprblue]{hyperref}

\def\confName{CVPR}
\def\confYear{2027}

\title{Renormaly: Reference-Anchored Renormalization for Anomaly Detection Under Distribution Shift}

\author{
Dat Cao$^{1*}$ \quad
Son Nghiem$^{1,2*}$ \quad
Phan Nguyen$^{1*}$ \quad
Jun Rekimoto$^{2,3}$ \quad
Jhih-Ciang Wu$^{4\dagger}$\\[2pt]
$^{1}$ Korea Advanced Institute of Science and Technology (KAIST), South Korea\\
$^{2}$ Sony Computer Science Laboratories (Sony CSL), Japan\\
$^{3}$ The University of Tokyo, Japan\\
$^{4}$ National Taiwan Normal University (NTNU), Taiwan
}

\newcommand{\authorfootnotes}{{%
  \renewcommand{\thefootnote}{\fnsymbol{footnote}}%
  \footnotetext[1]{Equal contribution.}%
  \footnotetext[2]{Corresponding author: \texttt{jcwu@csie.ntnu.edu.tw}}}}

\begin{document}
\maketitle
\begin{abstract}
Industrial anomaly detectors degrade substantially under imaging conditions that differ from training, such as changes in illumination, background, or viewpoint. While a common remedy is to learn shift-invariant features, we show that misalignment of the normal reference remains a key bottleneck: much of the lost accuracy can be recovered without modifying the features. We find that augmenting frozen kNN memory banks with target normals recovers at least 80\% of shift-induced I-AUROC loss across evaluated benchmarks and backbones. However, this recalibration requires target normals from each new setup, which may be scarce or unavailable. We propose \textbf{Renormaly}, which \textbf{Re}normalizes the residuals of a frozen detector against its source \textbf{Norm}al reference, without target-side training, in two deployment scenarios: source-only and few-shot. In the source-only setting, Renormaly estimates nuisance directions from content-preserving perturbations of source images and caps excess feature spread along these directions at the source-normal level, preserving localized defect evidence. In the few-shot setting, it re-centers the residuals on the source reference, using an empirical-Bayes weight that balances target evidence against sampling noise. Renormaly achieves {\em state-of-the-art performance under distribution shift} in both settings. In the few-shot setting, our method outperforms competing methods using half as many target normal samples, demonstrating its sample efficiency. Notably, this robustness does not come at the expense of conventional AD performance: Renormaly remains competitive on standard benchmarks that do not explicitly evaluate distribution shift, including VisA, Real-IAD, and MVTec-AD.

\end{abstract}

\section{Introduction}
\label{sec:intro}

\begin{figure}[t]
  \centering
  \includegraphics[width=0.48\textwidth]{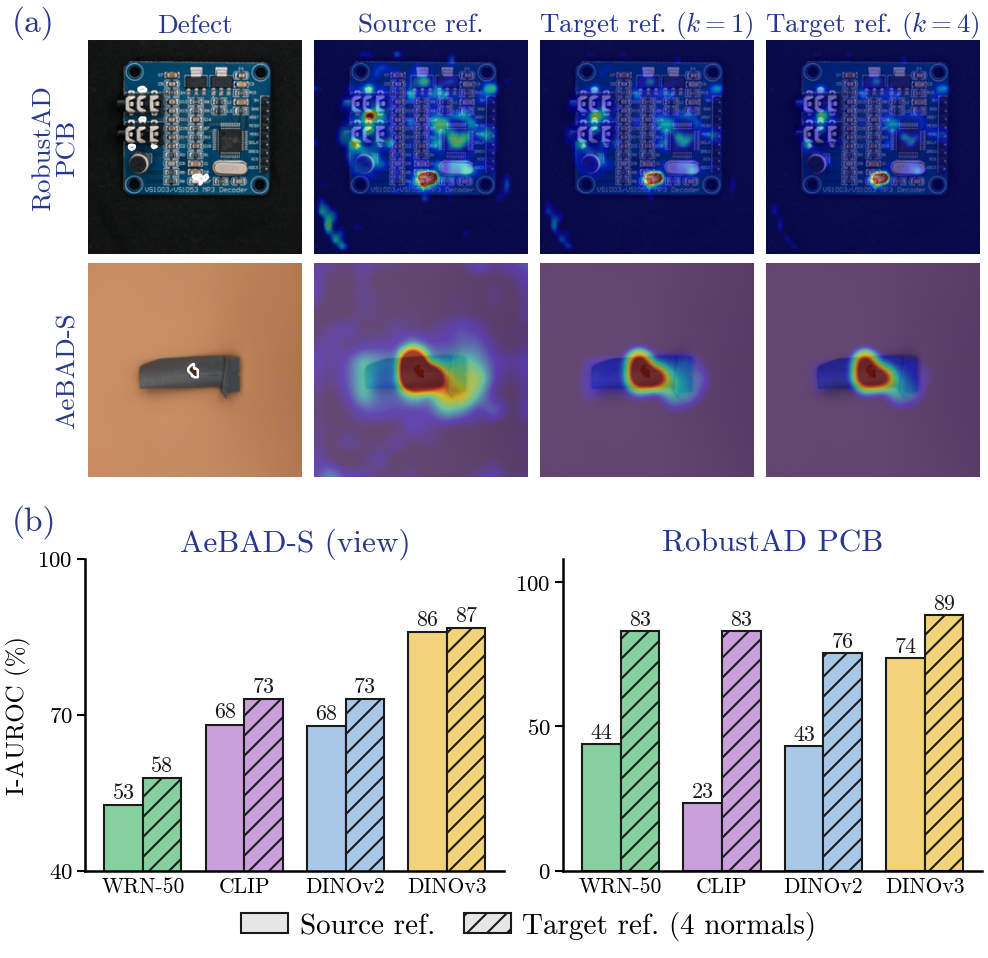}
  \caption{
  \textbf{Under acquisition shift, the normal reference is a primary bottleneck.} (a) Our method keeps the detector frozen and renormalizes its residuals with $k$ target normals: shift-induced false alarms disappear. (b) The bottleneck is consistent: for plain kNN on four frozen backbones, adding 4 target normals to the source memory bank recovers much of the image AUROC lost under shift.}
  \label{fig:teaser}
\end{figure}

Industrial anomaly detection (AD) has made remarkable progress in recent years. Rapid advances in visual backbones, feature representations, and anomaly modeling have pushed performance on established benchmarks such as MVTec-AD and VisA toward saturation~\cite{Dinomaly,RAD2026,MVTecAD,Spot}. These benchmarks support consistent comparison, but their relatively controlled acquisition conditions provide limited evidence of robustness to changes in the imaging setup. In deployment, changes in illumination, background, viewpoint, and camera configuration can shift normal samples away from the learned normal distribution and cause false anomaly responses. Despite its practical importance, anomaly detection under such distribution shifts remains comparatively underexplored.\authorfootnotes

Recent benchmarks are built to expose exactly this gap. AeBAD~\cite{AeBAD} and RobustAD~\cite{RobustAD} capture test images under acquisition conditions that differ from training, while synthetic corruptions of MVTec-AD~\cite{ADShift, FiCO} simulate such shifts in a controlled way. The substantial degradation of existing detectors motivates methods that are \textbf{robust to nuisance variations while remaining sensitive to defects}. Here, nuisance variations refer to acquisition changes that alter image appearance without changing defect status.

Robustness to such nuisance variation is particularly difficult for feature-based anomaly detection, which detects defects as deviations from normal feature representations \cite{RD4AD,UniAD}. When the acquisition conditions change, such a detector can fail for two distinct reasons. First, the representation itself may lose the information that separates defects from normal appearance, in which case no comparison can recover them. Second, the features may still separate defects, but the normal reference they are compared against no longer describes normal images of the new setup. This reference consists of normal statistics collected under the source setup, so the shift itself is read as a deviation. Invariant-representation methods address the first failure mode by relearning features to resist the shift \cite{ADShift, FiCO}. Reference mismatch, instead, calls for correcting the mismatch while keeping the detector frozen. Existing approaches address it implicitly by adding a few target normals to stored features \cite{WinCLIP,RAD2026} or fine-tuning on them \cite{INP-Former}, without explicitly estimating how far these samples should be trusted. We find that reference mismatch dominates. Across frozen detectors built on different backbones, augmenting the reference with only a few target normal images recovers most of the accuracy lost to the shift (\cref{fig:teaser}).

Based on these findings, we propose \textbf{Renormaly}, which renormalizes the residuals of a frozen detector against its source reference in two steps corresponding to source-only and few-shot deployment. In the source-only scenario, Renormaly estimates nuisance directions from content-preserving source perturbations and removes residual energy along them only when it exceeds normal training levels. With a few target normals, it re-centers the remaining residuals using an empirical-Bayes weight that balances the observed source–target offset against sampling noise. 

In summary, our contributions are:
\begin{itemize}
    \item We establish theoretically and empirically that normal-reference mismatch is the primary failure source under acquisition shift. By bounding information loss in frozen features, we show that this loss vanishes for shifts shared across target images (\cref{sec:diagnosis}), and recover most lost accuracy by correcting only the reference across backbones.
    \item We introduce Renormaly, a post-hoc reconstruction framework that corrects normal-reference mismatch in a frozen detector by renormalizing its residuals onto the source reference, using only source perturbations or a few target normals.
    \item Our method achieves \textbf{state-of-the-art} results under real acquisition shifts on AeBAD-S and RobustAD and, unlike prior methods designed for distribution shift, retains competitive accuracy on conventional AD benchmarks.
\end{itemize}

\section{Related Work}

\subsection{Feature-Based Industrial Anomaly Detection}
A common approach identifies anomalies as deviations from the learned representation of normal data.
Such \emph{feature-reconstruction methods} assume that a model trained only on normal data would reconstruct normal features
more accurately than anomalous ones: teacher--student approaches such as RD4AD~\cite{RD4AD} and EfficientAD~\cite{EfficientAD}
use the teacher--student discrepancy as the anomaly score, while MMR~\cite{AeBAD}, Dinomaly~\cite{Dinomaly} and
INP-Former~\cite{INP-Former} reconstruct normal features under masking, relaxed constraints or normal prototypes.   %
On the other hand, \emph{embedding-based methods} characterize normal feature structure directly, by comparing test
patches with stored normal embeddings~\cite{SPADE, PatchCore} or modeling their distribution~\cite{PaDiM}, while recent methods
apply this idea to frozen foundation-model features~\cite{AnomalyDINO, RAD2026}. However, both families score a test image against a normal reference built under the training setup, so when imaging
conditions change, the shift of normal images away from this reference is itself scored as anomalous.   

\subsection{Anomaly Detection under Distribution Shift}
Conventional anomaly detectors often degrade substantially when conditions change. A recent line of
\emph{distribution-shift methods} explored ways to make anomaly detection more robust: AeBAD~\cite{AeBAD} introduced a
benchmark where the test set differs from the training set in background, illumination and viewpoint, and its accompanying
MMR method improves robustness through masked multi-scale reconstruction.
RobustAD~\cite{RobustAD} similarly records real objects under controlled changes of acquisition conditions.   %
ADShift~\cite{ADShift} learns distribution-invariant normality from augmented out-of-distribution views, while
FiCo~\cite{FiCO} builds upon this idea by filtering and compensating for distribution-specific information within a
reverse-distillation framework. These methods alleviate distribution sensitivity through augmentation, invariant feature learning, or target-domain adaptation. However, they only partially address the problem, as the nuisance is absorbed into retrained representations, while the mismatch between test images and the normal reference is never explicitly corrected.

\subsection{Concept Erasure and Invariant Learning}
Concept erasure and invariant learning methods provide broader mechanisms for controlling nuisance information in learned
representations. INLP~\cite{NullItOut} iteratively projects representations onto the null spaces of linear concept classifiers,
whereas LEACE~\cite{LEACE} derives a closed-form linear erasure transformation. Domain-adversarial methods such as
DANN~\cite{DANN} suppress domain-predictive information, while self-supervised methods such as SimCLR~\cite{chen2020simclr} encourage
invariance across paired augmented views. MAST~\cite{MAST} further factorizes representations into augmentation-induced
subspaces, allowing different invariances to be represented separately. These approaches extend nuisance handling beyond dataset-specific adaptation. However, they typically modify backbone representations globally, either require concept or domain supervision or enforce invariance without considering how nuisance and task-relevant signals are spatially expressed.
Closer to our setting, batch-normalization adaptation keeps the weights and re-estimates feature statistics on the target, mixing them with source statistics through a fixed, hand-set weight~\cite{Schneider2020}.
In contrast, our method uses label-free, content-preserving interventions to estimate nuisance directions and correct reconstruction discrepancies while keeping the backbone and detector frozen. When target normals are available, their weight against the source is determined by the observed source–target deviation relative to sampling noise.

\section{A Closer Look at Failure under Acquisition Shift}
\label{sec:diagnosis}

Feature-based anomaly detection flags a defect when the features of a test image deviate from a \emph{normal reference}
built from normal source images. Under acquisition shift, however, normal images deviate as well, and the detector reads the
shift itself as a defect. Two explanations are possible, and they call for opposite remedies: either the features have lost
the information that separates defects, and the detector must be retrained, or they still carry it, and only the reference
is outdated. Unfortunately, the drop in detection performance cannot tell them apart: both causes lower the detector's ROC curve, and with it AUROC and every other summary of the curve. We separate the two, and find that the second dominates.

\textbf{Splitting the loss.}
Let $A_s$ and $A_t$ be the AUROC of a detector on the source and the shifted setup under the source reference, and
$A^*$ the AUROC of the best ROC curve of the frozen features, which no score exceeds at any false-positive rate. We denote by $A_t^{\mathrm{ref}}$ the AUROC on the
shifted setup once only the reference is updated with normal images of that setup. The loss then splits exactly into three parts:
\begin{equation}
\begin{aligned}
A_s - A_t ={}&
\underbrace{(A^*_s - A^*_t)}_{\text{information loss}}
+ \underbrace{(A_t^{\mathrm{ref}} - A_t)}_{\text{reference-recoverable}} \\
&+ \underbrace{\big[(A^*_t - A_t^{\mathrm{ref}}) - (A^*_s - A_s)\big]}_{\text{efficiency change}} .
\end{aligned}
\label{eq:split}
\end{equation}
Only the first part is beyond the reach of every detector. We report the second as the \emph{recovered fraction}
$\phi = (A_t^{\mathrm{ref}} - A_t)/(A_s - A_t)$.

\begin{figure}[t]
\centering
\begin{minipage}[c]{0.56\linewidth}
\centering
\scriptsize
\setlength{\tabcolsep}{2.4pt}
\begin{tabular}{@{}lccc@{}}
\toprule
 & AUROC & \multicolumn{2}{c}{Loss recovered $\phi$} \\
\cmidrule(l){3-4}
Backbone & lost & median & $\phi\ge0.5$ \\
\midrule
\mkDt\ DINOv3 & 19.9 & 0.95 &  9/9  \\
\mkDs\ DINOv2 & 39.8 & 0.93 &  9/9  \\
\mkCL\ CLIP   & 45.2 & 0.99 & 11/11 \\
\mkWR\ WRN-50 & 41.2 & 0.93 & 10/11 \\
\bottomrule
\end{tabular}

\end{minipage}\hfill
\begin{minipage}[c]{0.42\linewidth}
\centering
\includegraphics[width=\linewidth]{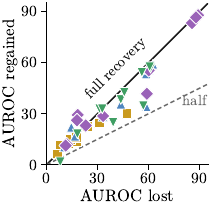}
\end{minipage}
\caption{\textbf{Rebuilding only the reference recovers most of the loss.} Frozen kNN detectors, with the reference rebuilt
from 16 normal images of each shifted RobustAD domain where the detector loses at least 5 AUROC points.
Left: mean AUROC lost, median recovered fraction $\phi$ and the number of affected domains with $\phi\ge0.5$. Right: one point per detector and domain,
on the diagonal when the loss is fully recovered.}
\label{fig:reref}
\end{figure}
\textbf{What a reference can and cannot recover.}
Not every shift destroys information. A shift shared by all images only moves the reference; information is lost only when
the nuisance varies from image to image along the direction that separates defects.
\begin{proposition}[Information limit]
\label{prop:info}
Let the nuisance be independent of the defect label and of the source features.
(i) The best ROC curve of the shifted setup lies on or below that of the source at every false-positive rate, so
$A^*_t \le A^*_s$. The two curves coincide when the shift acts as the same invertible map on all images.
(ii) Let the source features be Gaussian, $\mathcal{N}(\mu,\Sigma)$ for normal and $\mathcal{N}(\mu+a,\Sigma)$ for defective
images with offset $a\in\mathcal{A}$, and the nuisance additive, $n \sim \mathcal{N}(m,\Sigma_n)$. The two curves coincide if
and only if $\Sigma_n\Sigma^{-1}a = 0$ for all $a\in\mathcal{A}$, regardless of the mean $m$.
\end{proposition}
Part (i) is an instance of Blackwell's comparison of experiments~\cite{Blackwell1953}, and it bounds every summary of the
curve, such as the TPR at a fixed false-positive rate. Proofs are provided in the supplement. Since $A_t^{\mathrm{ref}} \le A^*_t$,~\eqref{eq:split} yields
\begin{equation}
A^*_s - A^*_t \;\le\; (A^*_s - A_s) + (1-\phi)\,(A_s - A_t).
\label{eq:bound}
\end{equation}

Therefore, a recovered fraction near one means that the shifted features have lost no more information than the detector already
leaves unused on the source.

\begin{figure*}[t]
  \centering
  \includegraphics[width=\textwidth]{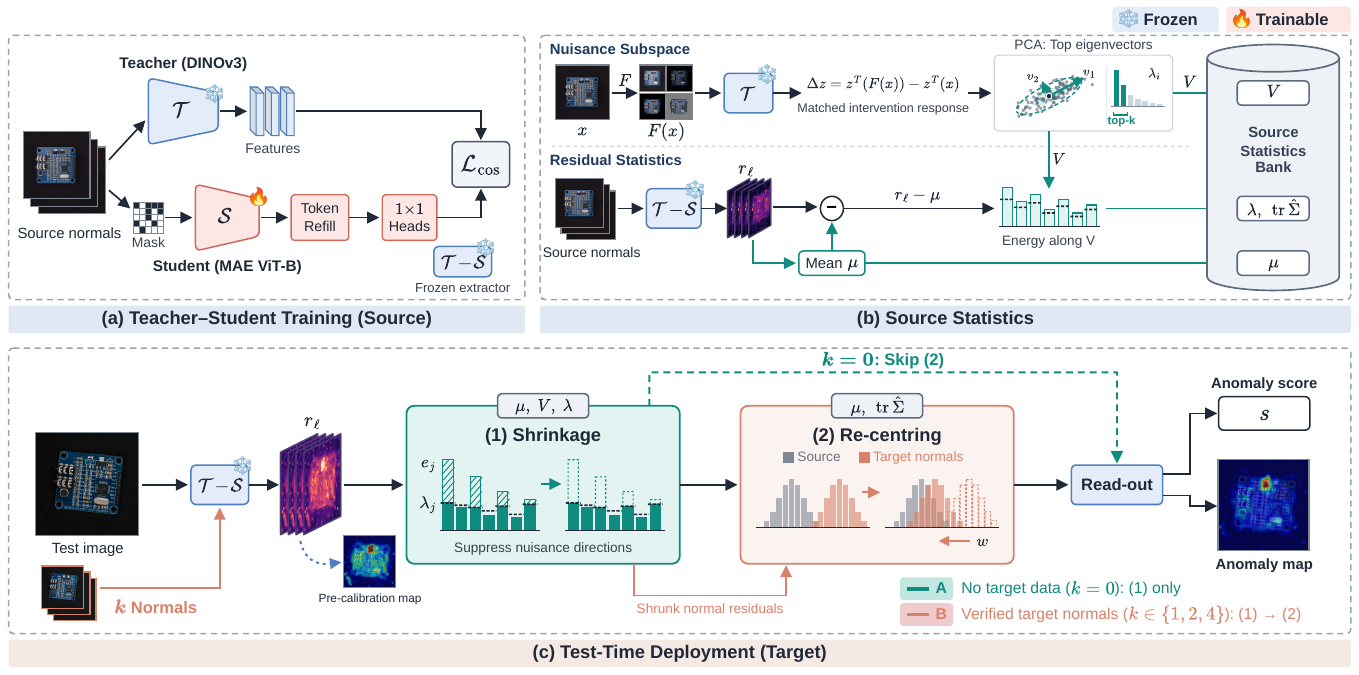}
   \caption{\textbf{Overview of Renormaly.}
(a)~The student \Stu{} is trained to reconstruct the teacher \Tch{} features from masked source normals, forming the frozen extractor \TS{} whose residual $r_\ell$ is scored.
(b)~From source normals alone, we estimate the nuisance subspace $V$ from matched intervention responses and the source reference statistics from the residuals.
(c)~At test time, (1)~\shrink{} caps each image's energy along $V$ at the source level, suppressing global nuisance while preserving local defects. (2)~Given $k$ verified target normals, \recenter{} aligns their mean to the source reference with a James--Stein weight discounting sampling noise. For $k{=}0$, step~(2) is skipped.}
  \label{fig:overview}
\end{figure*}

\textbf{Reference mismatch dominates.}
We measure $\phi$ for frozen kNN detectors on four backbones (\cref{fig:reref}). For each shifted RobustAD domain where a
detector clearly degrades, we rebuild only its reference from 16 normal images of that domain. The source reference receives 16 extra
normal images as well, so extra data alone cannot explain the gain. However large the loss, nearly all of it is recovered: the
recovered fraction stays close to one for every backbone, and falls below one half only once. 
Four target normals already show
the same effect (\cref{fig:teaser}b).
Wherever the detectors are accurate on the source,~\cref{eq:bound} then implies that their features have lost little
information under these shifts: the bottleneck is the reference.

\textbf{What deployment leaves open.}
The remedy is thus to realign test images with the reference while keeping the detector frozen.
\cref{prop:info}(i) also marks the limit of any such correction, including ours: any realignment acts only on how the test image’s frozen features are scored, so it can bring a detector closer to the best accuracy its
features allow, but cannot recover information they have lost. The rebuild above, however, relied on 16 verified normals of
every new setup, which deployment rarely provides. Two questions remain, and the next section answers them in turn.
Can test residuals be brought back to the reference before a single normal of the new setup has been seen? And when only a handful of target normals are available, their deviation from the source mixes the shift with their own sampling noise: can the data decide how far to trust them, without a hand-set weight?

\section{Method}
\label{sec:method}

\textit{Renormaly} answers both questions without retraining. It uses a detector whose reference is explicit
(\cref{sec:detector}), learns from the source alone along which directions a shift pulls residuals away from this reference (\cref{sec:prior}), and re-centers the residuals of the new setup on it using a few target normals, weighted against their sampling noise (\cref{sec:eb}).
\cref{fig:overview} gives an overview of our method.

\subsection{Expose the Reference}
\label{sec:detector}

A detector scores a test image against a reference of normal source data, and renormalizing against it requires an explicit reference to anchor to. A memory bank keeps it as a set of patches, with no single reference vector per token location. A
teacher--student detector reduces it to one vector per token: the training mean $\mu_\ell(p)$ of the residual
$r_\ell(x,p)=z^T_\ell(x,p)-z^S_\ell(x,p)$ at token position $p\in\{1,\dots,P\}$, near which normal images stay and from which
defects depart., near which normal images stay and from which defects depart. We use a frozen
DINOv3~\cite{DINOv3} ViT-B/16 teacher with blocks $\ell\in\mathcal{B}=\{2,4,6,8\}$ and an MAE~\cite{MAE} ViT-B student,
trained on source normals to reconstruct the teacher features $z^T_\ell(x,p)\in\mathbb{R}^C$ from a masked view. With the residual corrected by $\Delta$ (\cref{sec:prior,sec:eb}), each position is scored by
\begin{equation}
a(p)=\frac{1}{|\mathcal{B}|}\sum_{\ell\in\mathcal{B}}\big[1-\cos\big(z^T_\ell(p)-\Delta_\ell(p),\,z^S_\ell(p)\big)\big],
\label{eq:map}
\end{equation}
which is smoothed and z-scored with source statistics, and the top-1\% mean gives the image score.

\subsection{Anticipate the Shift}
\label{sec:prior}

Before the first target image arrives, the shift cannot be observed, but it can be anticipated: illumination, camera
response and background can be simulated on source images. The challenge is to isolate their effect on the features and remove it without suppressing defect evidence.

\textbf{An image is its own control.}
We compare each source normal image $x \in \mathcal{D}$ with content-preserving interventions of itself: photometric ones $\mathcal{F}_p$ (brightness,
contrast, gamma, color temperature, shading) mimic illumination and camera changes, and background ones $\mathcal{F}_b$,
which refill the region outside the foreground, mimic scene changes. The \emph{matched intervention response} (MIR)
\begin{equation}
\Delta z^T_\ell(x,F,p)=z^T_\ell(F(x),p)-z^T_\ell(x,p)
\label{eq:mir}
\end{equation}
cancels the content shared by $x$ and $F(x)$ and isolates the nuisance response to first order. The nuisance directions are
those along which these responses concentrate: per block $\ell$ and family $\mathcal{F}$, the top $d=16$ eigenvectors
$U^{\mathcal{F}}_\ell$ of their uncentered second moment
\begin{equation}
S^{\mathcal{F}}_\ell=\frac{1}{|\mathcal{D}||\mathcal{F}|P}\sum_{x,F,p}\Delta z^T_\ell(x,F,p)\,\Delta z^T_\ell(x,F,p)^\top,
\label{eq:mirmom}
\end{equation}
and $V_\ell=\mathrm{orth}([U^p_\ell,U^b_\ell])\in\mathbb{R}^{C\times 2d}$ spans both families.

\textbf{Shifts spread, defects don't.}
Projecting out these directions would also remove defect evidence aligned
with them. The key distinction is spatial extent: a global shift perturbs
the coordinate along $v_{\ell j}$ at nearly every token, whereas a local
defect affects only the tokens it covers.
Let $c_\ell(x,p)=V_\ell^\top(r_\ell(x,p)-\mu_\ell(p))$ denote the projected
residual coordinates. We define the image-level energy along $v_{\ell j}$
as
$e_{\ell j}(x)=\big(\mathrm{med}_p|c_{\ell j}(x,p)|\big)^2$.\footnote{For zero-mean Gaussian
coordinates, $e_{\ell j}$ equals the variance up to the constant
$\Phi^{-1}(3/4)^2$; since Eq.~\eqref{eq:prior} depends only on
$e_{\ell j}/\lambda_{\ell j}$, the constant is omitted.}
This median is largely insensitive to local defects but increases under
global shifts, whether through a change in spread or center. We set the
source level $\lambda_{\ell j}$ to the median of $e_{\ell j}$ across
training images, and shrink each coordinate using a single image-level
factor:
\begin{equation}
\Delta^{0}_\ell(x,p)=\sum_j
\Big(1-\sqrt{\tfrac{\lambda_{\ell j}}
{\max(e_{\ell j}(x),\lambda_{\ell j})}}\Big)
c_{\ell j}(x,p)\,v_{\ell j}.
\label{eq:prior}
\end{equation}

The factor is shared across all tokens, preserving their relative
coordinates along each direction. Thus, local defect contrast is retained,
while the image-level energy is reduced to $\lambda_{\ell j}$ whenever
it exceeds the source level.

\begin{proposition}[Minimal energy cap]
\label{prop:cap}
Among per-image rescalings of each coordinate by a factor in $[0,1]$, $\Delta^0$ is the smallest at every token that caps
the robust energy at $\lambda_{\ell j}$. It is zero when $e_{\ell j}(x)\le\lambda_{\ell j}$ for all $j$, and a defect on
fewer than half of the tokens can raise $e_{\ell j}(x)$ only to a quantile of the other tokens.
\end{proposition}
Shifts outside $\mathrm{span}\,V_\ell$, such as a change of viewpoint, pass unchanged and call for images of the new setup.

\begin{table*}[t]
\centering
\scriptsize\setlength{\tabcolsep}{5.2pt}\renewcommand{\arraystretch}{1.08}\setlength{\aboverulesep}{3pt}\setlength{\belowrulesep}{1.5pt}
\caption{\textbf{Comparison of anomaly detection and localization performance on the distribution-shift benchmarks AeBAD-S~\cite{AeBAD} and RobustAD~\cite{RobustAD}, with source-only $(k{=}0)$ and verified target normal settings.} AUPRO up to FPR 0.3. The highest of each column is shown in \textbf{bold}, and the second-highest is marked with \underline{underline}.}
\label{tab:main}
\begin{tabular}{@{}clc ccccccc ccccccc@{}}
\toprule
 & & & \multicolumn{3}{c}{\textbf{AeBAD-S \textit{(image-level)} }} & \multicolumn{4}{c}{\textbf{AeBAD-S \textit{(pixel-level)}} } & \multicolumn{3}{c}{ \textbf{RobustAD \textit{(image-level)}}} & \multicolumn{4}{c}{\textbf{RobustAD \textit{(pixel-level)}}} \\
\cmidrule(lr){4-6}\cmidrule(lr){7-10}\cmidrule(lr){11-13}\cmidrule(lr){14-17}
\textbf{\# k} & \textbf{Method} & \textbf{Venue} & AUROC & AP & F1 & AUROC & AP & F1 & AUPRO & AUROC & AP & F1 & AUROC & AP & F1 & AUPRO \\
\midrule
\multirow{7}{*}{\rotatebox{90}{$k{=}0$}} & PatchCore~\cite{PatchCore} & CVPR 2022 & 71.1 & 86.5 & 85.9 & \underline{93.4} & \textbf{13.0} & \textbf{20.2} & 85.0 & 70.8 & 67.1 & 68.8 & 91.1 & 8.0 & 15.4 & 71.1 \\
 & ADShift~\cite{ADShift} & ICCV 2023 & 74.7 & 88.9 & 86.3 & 91.1 & 10.2 & 15.2 & 89.0 & 71.3 & 69.7 & 72.1 & 89.5 & 6.8 & 12.0 & 67.3 \\
 & FiCo~\cite{FiCO} & AAAI 2025 & 60.1 & 79.8 & 84.6 & 88.8 & 4.3 & 9.1 & 74.2 & 68.9 & 66.7 & 71.0 & 90.4 & 6.0 & 11.7 & 69.4 \\
 & INP-Former~\cite{INP-Former} & CVPR 2025 & 73.6 & 87.0 & 88.4 & 90.4 & 9.7 & 16.4 & 85.8 & \underline{74.6} & \underline{71.3} & \underline{74.7} & \underline{95.5} & 11.4 & \underline{21.4} & \underline{83.5} \\
 & Dinomaly~\cite{Dinomaly} & CVPR 2025 & \underline{79.5} & \underline{90.1} & \underline{89.7} & 90.9 & 10.8 & 17.6 & 86.8 & 73.4 & 68.2 & 74.3 & 95.1 & 7.4 & 15.8 & 83.1 \\
 & RAD~\cite{RAD2026} & ICML 2026 & 74.5 & 88.0 & 87.1 & 92.6 & \underline{12.8} & 16.9 & \underline{90.6} & 73.9 & 70.8 & 74.0 & 95.1 & \underline{12.2} & 20.2 & 81.8 \\
 \cmidrule(l){2-17}
\rowcolor{oursbg} & \textbf{
Renormaly (Ours)} & -- & \textbf{87.6} & \textbf{95.6} & \textbf{90.5} & \textbf{93.8} & 11.7 & \underline{20.0} & \textbf{91.0} & \textbf{85.4} & \textbf{80.7} & \textbf{77.9} & \textbf{98.4} & \textbf{16.9} & \textbf{25.8} & \textbf{92.8} \\
\midrule
\multirow{7}{*}{\rotatebox{90}{$k{=}1$}} & PatchCore~\cite{PatchCore} & CVPR 2022 & 71.9 & 86.8 & 86.1 & \underline{93.6} & \textbf{13.0} & \textbf{20.2} & 85.1 & 78.6 & 73.7 & 73.3 & 92.9 & 8.8 & 16.4 & 75.6 \\
 & ADShift~\cite{ADShift} & ICCV 2023 & 78.3 & 90.5 & 87.7 & 91.3 & 10.4 & 15.0 & 89.1 & 81.9 & 76.9 & 77.1 & 96.0 & 9.7 & 19.1 & 84.9 \\
 & FiCo~\cite{FiCO} & AAAI 2025 & 76.2 & 87.8 & 88.2 & 91.5 & 9.2 & 13.8 & 88.2 & \underline{83.0} & \underline{78.3} & \underline{77.7} & 96.2 & 8.8 & 18.2 & 85.7 \\
 & INP-Former~\cite{INP-Former} & CVPR 2025 & 78.6 & 89.8 & 89.2 & 90.4 & 9.1 & 15.2 & 86.0 & 81.8 & 74.9 & 76.8 & 96.3 & 12.7 & \underline{22.9} & 86.6 \\
 & Dinomaly~\cite{Dinomaly} & CVPR 2025 & \underline{83.0} & \underline{92.2} & \underline{89.5} & 90.4 & 9.2 & 15.5 & 86.5 & 80.9 & 72.4 & 77.6 & \underline{96.9} & 10.2 & 19.5 & \underline{88.7} \\
 & RAD~\cite{RAD2026} & ICML 2026 & 74.8 & 88.2 & 87.2 & 92.7 & \underline{12.9} & 16.9 & \underline{90.7} & 80.5 & 74.7 & 76.5 & 96.3 & \underline{14.1} & 22.1 & 85.5 \\
 \cmidrule(l){2-17}
\rowcolor{oursbg} & \textbf{Renormaly (Ours)} & -- & \textbf{90.2} & \textbf{96.7} & \textbf{91.5} & \textbf{94.1} & 12.8 & \underline{20.1} & \textbf{91.3} & \textbf{88.2} & \textbf{83.3} & \textbf{80.0} & \textbf{98.5} & \textbf{17.3} & \textbf{26.2} & \textbf{93.3} \\
\midrule
\multirow{7}{*}{\rotatebox{90}{$k{=}2$}} & PatchCore~\cite{PatchCore} & CVPR 2022 & 72.2 & 87.0 & 86.3 & \underline{93.6} & \underline{13.1} & \underline{20.2} & 85.0 & 82.1 & 77.2 & 75.8 & 94.3 & 9.6 & 17.5 & 79.7 \\
 & ADShift~\cite{ADShift} & ICCV 2023 & 79.1 & 90.8 & 88.0 & 91.5 & 10.6 & 15.1 & 89.1 & 84.6 & 79.7 & 79.0 & 96.8 & 11.6 & 20.9 & 87.5 \\
 & FiCo~\cite{FiCO} & AAAI 2025 & 77.4 & 88.8 & 88.5 & 91.6 & 9.7 & 14.0 & 88.3 & 85.5 & \underline{80.7} & 79.3 & 97.2 & 10.8 & 19.8 & 89.0 \\
 & INP-Former~\cite{INP-Former} & CVPR 2025 & 80.1 & 90.4 & 89.3 & 90.8 & 9.5 & 15.8 & 86.5 & 84.4 & 78.1 & 78.6 & 96.8 & 14.1 & 24.5 & 88.4 \\
 & Dinomaly~\cite{Dinomaly} & CVPR 2025 & \underline{84.5} & \underline{92.8} & \underline{90.1} & 91.3 & 10.0 & 16.7 & 87.4 & 85.0 & 77.5 & \underline{80.1} & \underline{97.6} & 11.7 & 21.2 & \underline{90.9} \\
 & RAD~\cite{RAD2026} & ICML 2026 & 75.2 & 88.3 & 87.3 & 92.8 & 13.0 & 16.9 & \underline{90.8} & \underline{85.6} & 79.8 & 79.5 & \underline{97.6} & \underline{15.9} & \underline{24.6} & 89.6 \\
 \cmidrule(l){2-17}
\rowcolor{oursbg} & \textbf{Renormaly (Ours)} & -- & \textbf{91.6} & \textbf{96.6} & \textbf{91.6} & \textbf{94.1} & \textbf{13.2} & \textbf{20.3} & \textbf{91.3} & \textbf{89.3} & \textbf{84.6} & \textbf{81.6} & \textbf{98.6} & \textbf{17.7} & \textbf{26.6} & \textbf{93.8} \\
\midrule
\multirow{7}{*}{\rotatebox{90}{$k{=}4$}} & PatchCore~\cite{PatchCore} & CVPR 2022 & 73.1 & 87.4 & 86.4 & \underline{93.8} & \underline{13.1} & \underline{20.2} & 85.0 & 84.8 & 79.5 & 78.1 & 95.0 & 10.0 & 18.0 & 82.1 \\
 & ADShift~\cite{ADShift} & ICCV 2023 & 80.2 & 91.2 & 88.3 & 91.7 & 10.8 & 15.2 & 89.3 & 85.8 & 81.1 & 79.9 & 97.1 & 12.5 & 21.7 & 88.4 \\
 & FiCo~\cite{FiCO} & AAAI 2025 & 79.4 & 90.2 & 88.6 & 91.8 & 10.4 & 14.6 & 88.6 & 86.6 & 82.0 & 80.3 & 97.5 & 11.8 & 21.0 & 90.0 \\
 & INP-Former~\cite{INP-Former} & CVPR 2025 & 83.9 & 92.1 & 90.4 & 91.2 & 9.6 & 16.2 & 86.9 & 86.7 & 80.8 & 80.0 & 97.0 & 15.0 & 25.8 & 89.1 \\
 & Dinomaly~\cite{Dinomaly} & CVPR 2025 & \underline{87.3} & \underline{94.1} & \underline{91.1} & 91.8 & 10.1 & 16.7 & 88.0 & 88.5 & 81.6 & \underline{82.3} & 97.9 & 13.2 & 23.1 & \underline{91.8} \\
 & RAD~\cite{RAD2026} & ICML 2026 & 76.0 & 88.6 & 87.5 & 92.9 & \underline{13.1} & 16.8 & \underline{90.9} & \underline{88.7} & \underline{83.2} & \underline{82.3} & \underline{98.0} & \underline{17.2} & \underline{26.2} & 91.2 \\
 \cmidrule(l){2-17}
\rowcolor{oursbg} & \textbf{Renormaly (Ours)} & -- & \textbf{91.7} & \textbf{96.7} & \textbf{91.7} & \textbf{94.1} & \textbf{13.2} & \textbf{20.3} & \textbf{91.3} & \textbf{91.3} & \textbf{86.7} & \textbf{83.8} & \textbf{98.8} & \textbf{18.4} & \textbf{27.4} & \textbf{94.5} \\
\bottomrule
\end{tabular}
\end{table*}

\subsection{Let the Data Weigh the Normals}
\label{sec:eb}

Verified normals $x_1,\dots,x_k$ of the new setup reveal what the source could not anticipate. They act on the corrected
residuals $\tilde r=r-\Delta^0$, so they only explain what \cref{sec:prior} leaves, and \textit{Renormaly} reduces to
Eq.~\eqref{eq:prior} at $k=0$.

\textbf{Follow the normals.}
Per token and block, let $m$ be the mean of the normals' residuals $\tilde r$ and $U$ an orthonormal basis of their centered
span, of rank at most $k-1$. The target term
\begin{equation}
\Delta^{\mathrm{tgt}}_\ell=(m_\ell-\mu_\ell)+U_\ell U_\ell^\top(\tilde r_\ell-m_\ell)
\label{eq:tgt}
\end{equation}
maps a residual to $(I-UU^\top)(\tilde r-m)+\mu$: the normal's mean $m$ is carried onto $\mu$, and the variation shared by the normals no
longer counts as a deviation.

\textbf{Shift or noise?}
With one to four normals, the offset $m-\mu$ in Eq.~\eqref{eq:tgt} contains both the shift and the sampling noise of $m$.
Subtracting it in full fits the reference to the $k$ normals and flags normal test images that differ from them by chance;
omitting it leaves the shift uncorrected. Since $\mathbb{E}\|m-\mu\|^2=\|\theta-\mu\|^2+\mathrm{tr}\,\Sigma/k$ for normals with
mean $\theta$ and covariance $\Sigma$, the noise share can be estimated from the normals themselves. Removing it gives, per block,
the positive-part James--Stein weight~\cite{james1961estimation, Efron01031973}, the estimated fraction of the offset energy that is shift:
\begin{equation}
\begin{gathered}
w_\ell=\Big[\,1-\frac{\sum_p \mathrm{tr}\,\hat\Sigma_\ell(p)/k}{\sum_p\|m_\ell(p)-\mu_\ell(p)\|^2}\,\Big]_+ ,\\
\Delta_\ell=\Delta^0_\ell+w_\ell\,\Delta^{\mathrm{tgt}}_\ell ,
\end{gathered}
\label{eq:eb}
\end{equation}
with $\hat\Sigma$ the sample covariance of the normals ($k\ge2$) or of the corrected training residuals ($k=1$). The weight
stays near zero within the noise and approaches one as the shift or $k$ grows.
\begin{proposition}
\label{prop:js}
If the corrected residuals of $k\ge2$ normals are i.i.d.\ Gaussian with covariance $\Sigma$ in a block, $w(m-\mu)$ has a
lower expected squared error than $m-\mu$ for every target mean whenever
$(\mathrm{tr}\,\Sigma)^2\ge 4\lambda_{\max}(\Sigma)\,\mathrm{tr}\,\Sigma+2\,\mathrm{tr}(\Sigma^2)/(k-1)$.
\end{proposition}
It suffices that $\mathrm{tr}\,\Sigma/\lambda_{\max}(\Sigma)\ge 4+2/(k-1)$, \ie, that the residuals spread over more than a
few directions. The proof is provided in the supplementary material.

\newcommand{\pub}[1]{{\footnotesize\itshape\scshape\oldstylenums{#1}}}
\section{Experiments}
\subsection{Experimental Settings}
\noindent\textbf{Baselines.}
We compare with six detectors from three families. \emph{Memory banks} score a test patch by its distance to stored
normal patches: PatchCore~\cite{PatchCore} and RAD~\cite{RAD2026}. \emph{Learned normality models} train a network on source normals: Dinomaly~\cite{Dinomaly}, which reconstructs foundation
features, and INP-Former~\cite{INP-Former}, which extracts normal prototypes from the test image itself.
\emph{Distribution-shift-specific methods} are designed for distribution shift: ADShift~\cite{ADShift} and
FiCo~\cite{FiCO} learn shift-invariant features from augmented source images and align the test features at
inference. Every baseline uses its official code and default single-class configuration (backbone, resolution and
schedule).

\noindent\textbf{Datasets.}
We evaluate robustness on two benchmarks with real acquisition shift. AeBAD-S~\cite{AeBAD} images aero-engine blades in one unshifted
and three shifted test domains (background, illumination and view). RobustAD~\cite{RobustAD} has three categories,
MetalParts, PCB and PiledBags, each with an unshifted first domain and five or six domains captured under different real
conditions. Together they give $23$ domains, of which $19$ are distribution-shifted. PiledBags is defined for
image-level classification only. MVTec-AD~\cite{MVTecAD}, VisA~\cite{Spot}, and Real-IAD~\cite{RealIAD} at $k=0$ measure performance on conventional AD datasets. Per-domain statistics are provided in the supplementary material.

\noindent\textbf{Metrics and Protocol.}
We report image-level AUROC, AP and F1-max for anomaly detection, and pixel-level AUROC, AP, F1-max, and AUPRO (up to FPR 0.3) for anomaly localization. Every detector is trained once per category on source normals. In each target domain we repeat $5$ random draws: each
draw sets aside $16$ normal test images, the $k\in\{0,1,2,4\}$ verified normals are the first $k$ of them, and all methods are scored on the same remaining images. Results are averaged over the $5$ draws and then over
domains. 

\subsection{Comparison with SOTAs on real-world domain-shift benchmarks}
\label{sec:sota}
\cref{tab:main} compares \textit{Renormaly} with six baseline detectors. Our method ranks first on every image metric at every
$k$ on both benchmarks. The margin is largest when no target normal is available: at $k=0$, Renormaly exceeds the best
baseline in I-AUROC by \textcolor{axisred}{$8.1{\uparrow}$} points on AeBAD-S (Dinomaly~\cite{Dinomaly}) and by \textcolor{axisred}{$10.8{\uparrow}$} on RobustAD (INP-Former~\cite{INP-Former}).

\noindent\textbf{Fewer normals for the same accuracy.}
Every baseline improves as target normals arrive. On RobustAD they gain $12$ to $18$ I-AUROC points from $k=0$ to $k=4$,
consistent with the diagnosis of \cref{sec:diagnosis} that most of the loss lies in the reference. \textit{Renormaly} needs fewer
normals. On RobustAD, with two normals it exceeds every baseline with four (\textcolor{axisblue}{$89.3$} vs.\  \textcolor{axisblue}{$88.7$}). On AeBAD-S, the source prior alone at $k{=}0$ (\textcolor{axisblue}{$87.6$}) exceeds every baseline at $k{=}4$ (\textcolor{axisblue}{$87.3$}), and additional target normal images further improve its accuracy.

\noindent\textbf{Localization.}
\textit{Renormaly} gives the best P-AUROC and AUPRO on both benchmarks and the best value of every pixel metric on RobustAD.
At $k=0$, Renormaly improves P-AUROC by
\textcolor{axisred}{$2.9{\uparrow}$} points and AUPRO by
\textcolor{axisred}{$9.3{\uparrow}$} points over the strongest
baseline on RobustAD.
Even with $k=4$, these advantages persist, with gains of
\textcolor{axisred}{$0.8{\uparrow}$} and
\textcolor{axisred}{$2.7{\uparrow}$} points, respectively. On AeBAD-S, PatchCore reaches a higher P-AP and P-F1 at $k = 0$ and $k=1$. However, as the number of verified target normal images increases, our method regains the advantage. AUPRO
weighs every defect region equally, whereas P-AP is dominated by the largest defects, which suggests that \textit{Renormaly} finds more of the defects
while the memory banks outline the large ones more precisely.

\subsection{Comparison with SOTAs on conventional AD benchmarks}

\label{sec:std}
Standard benchmarks carry no acquisition shift, so they can answer whether handling shift costs accuracy on clean data.
\cref{tab:conv} reports our single-class results at $k=0$ on VisA~\cite{Spot}, MVTec-AD~\cite{MVTecAD} and Real-IAD~\cite{RealIAD}.

\noindent\textbf{Competitive without shift.}
On VisA, \textit{Renormaly} ties Dinomaly for the best image AUROC (\textcolor{axisblue}{$98.9$}) and has the best AUPRO (\textcolor{axisblue}{$95.8$}). On the large-scale,
multi-view Real-IAD, it is best on all three metrics (\textcolor{axisblue}{$92.2/99.3/97.0$}). MVTec-AD is close to saturation: every method
lies between $97.0$ and $99.7$ image AUROC, and Renormaly (\textcolor{axisblue}{$99.2$}) is \textcolor{axisred}{$0.5{\downarrow}$} points below the best.

\noindent\textbf{Clean rankings do not survive shift.}
The methods that lead on MVTec-AD do not stay ahead once the setup changes. Dinomaly and INP-Former trail Renormaly by
\textcolor{axisred}{$8.1{\downarrow}$} and \textcolor{axisred}{$14.0{\downarrow}$} image AUROC points on AeBAD-S, and by \textcolor{axisred}{$12.0{\downarrow}$} and \textcolor{axisred}{$10.8{\downarrow}$} on RobustAD (\cref{tab:main}, $k=0$).

\begin{table}[t]
\centering
\scriptsize
\setlength{\tabcolsep}{2.7pt}
\renewcommand{\arraystretch}{1.08}
\setlength{\aboverulesep}{1.5pt}
\setlength{\belowrulesep}{1.5pt}
\caption{\textbf{Comparison on standard unsupervised AD benchmarks under single-class setting at $k{=}0$}.
I-AU: Image AUROC,  P-AU: Pixel AUROC, PRO: AUPRO. \textbf{Bold:} best result, \underline{underline:} the second highest result.}  
\label{tab:conv}
\begin{tabular}{@{}l ccc ccc ccc@{}}
\toprule
 & \multicolumn{3}{c}{\textbf{VisA~\cite{Spot}}}
 & \multicolumn{3}{c}{\textbf{MVTec-AD~\cite{MVTecAD}}}
 & \multicolumn{3}{c}{\textbf{Real-IAD~\cite{RealIAD}}} \\
\cmidrule(lr){2-4}
\cmidrule(lr){5-7}
\cmidrule(l){8-10}
Method & I-AU & P-AU & PRO & I-AU & P-AU & PRO & I-AU& P-AU & PRO \\
\midrule

PatchCore \pub{cvpr'2022}
& 94.6 & 98.5 & 92.6
& 99.1 & 98.1 & 93.0
& 90.5 & -- & 94.0 \\

ADShift \pub{iccv'2023}
& 94.6 & 98.3 & 90.3
& 97.0 & 96.9 & 91.9
& -- & -- & -- \\

FiCo \pub{aaai'2025}
& 88.9 & \underline{98.9} & 93.3
& 97.7 & 97.7 & 93.3
& 89.1 & 98.9 & 93.1 \\

INP-Former \pub{cvpr'2025}
& \underline{98.5} & 98.5 & 94.8
& \underline{99.6} & 98.3 & \textbf{95.6}
& \underline{92.1} & -- & 95.6 \\

Dinomaly \pub{cvpr'2025}
& \textbf{98.9} & \underline{98.9} & \underline{95.1}
& \textbf{99.7} & \textbf{99.9} & \underline{95.0}
& 92.0 & \underline{99.1} & 95.1 \\

RAD \pub{icml'2026}
& 97.5 & \textbf{99.1} & 94.6
& \underline{99.6} & \underline{98.5} & 94.9
& 92.0 & \underline{99.1} & \underline{96.1} \\

\midrule
\rowcolor{oursbg}
\textbf{Renormaly (Ours)}
& \textbf{98.9} & 98.8 & \textbf{95.8}
& 99.2 & 98.1 & 93.7
& \textbf{92.2} & \textbf{99.3} & \textbf{97.0} \\
\bottomrule
\end{tabular}
\end{table}

\begin{figure}[!tb]
  \centering
  \includegraphics[width=\linewidth]{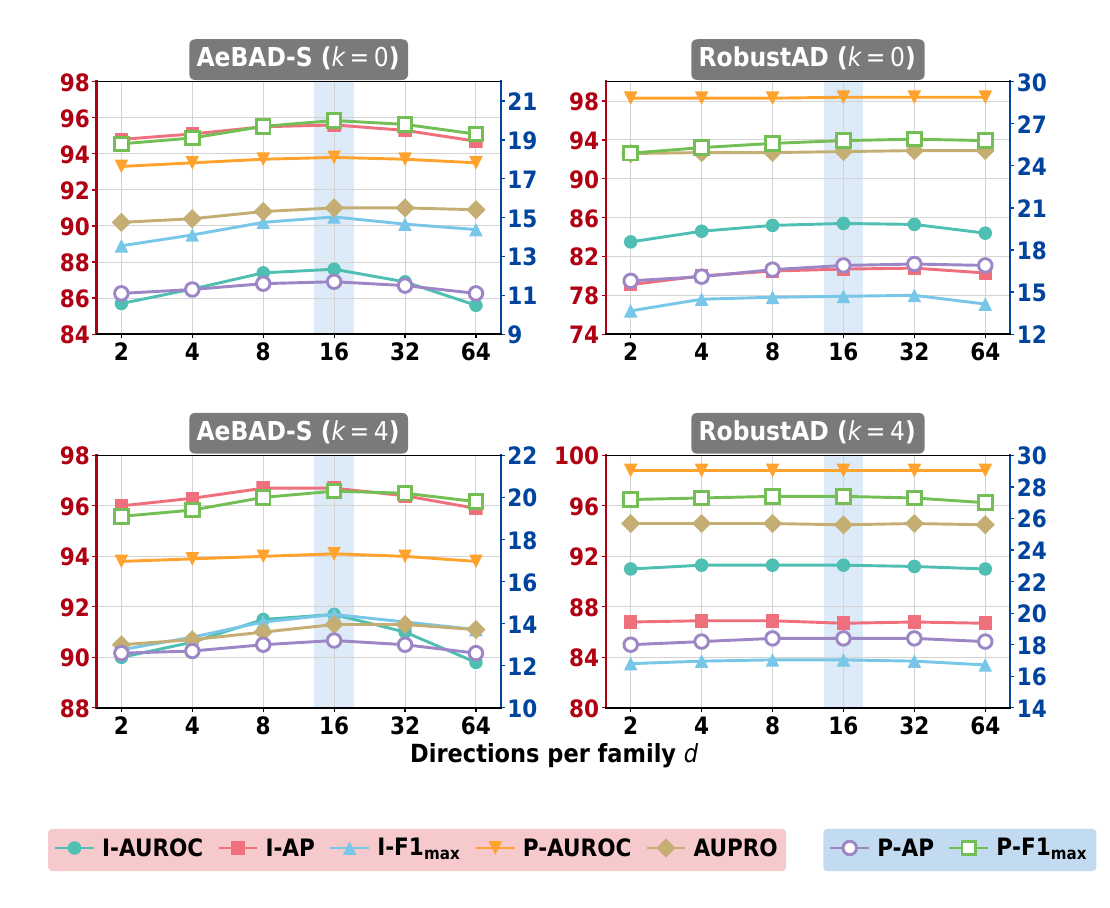}
 \caption{\textbf{Sensitivity to the number $d$ of nuisance directions per perturbation family at $k{=}0$ (source-only) and $k{=} 4$}.
Pixel-level AP and F1-max use the \textcolor{axisblue}{right vertical axis}, while the other metrics share the
\textcolor{axisred}{left vertical axis}. \textbf{Performance is stable}, varying by at most $0.7$ points for $d$ between $8$ and $32$,
and $d{=}16$ (our configuration) is within $0.2$ points of the best $d$ result.}
\label{fig:ranksensitivity}

\end{figure}

\subsection{Ablation study}

\begin{table*}[t]
\centering
\scriptsize\setlength{\tabcolsep}{2.6pt}\renewcommand{\arraystretch}{1.1}\setlength{\aboverulesep}{1.5pt}\setlength{\belowrulesep}{1.5pt}
\caption{\textbf{Component and design ablation (I-AUROC/AUPRO)}. (a) Source-only prior ($k=0$): each row changes one design choice of \cref{sec:prior}. (b) Target term ($k\ge1$ verified normals): each row changes one choice of
\cref{sec:eb}. Best results are shown in \textbf{bold}.}
\label{tab:abl}
\begin{minipage}[t]{0.305\linewidth}\centering
\textbf{(a)} Source-only prior, $k=0$\\[2pt]
\begin{tabular}{@{}l|cc@{}}
\toprule
& \textbf{AeBAD-S~\cite{AeBAD}} & \textbf{RobustAD~\cite{RobustAD}} \\
\midrule
Prior usage $\downarrow$ 
& \multicolumn{2}{c}{$k{=}0$} \\
\midrule
none (frozen detector) & 84.1/88.5 & 83.0/90.3 \\
random directions & 85.3/89.0 & 83.9/92.4 \\
PCA directions & 86.1/90.2 & 85.0/92.4 \\
project out $V$ & 86.5/90.1 & 83.6/92.0 \\
\rowcolor{oursbg}
\textbf{MIR, energy cap} & \textbf{87.6/91.0} & \textbf{85.4/92.8} \\
\bottomrule
\end{tabular}
\end{minipage}\hfill
\begin{minipage}[t]{0.675\linewidth}
\centering
\scriptsize\setlength{\tabcolsep}{5.0pt}

\textbf{(b)} Target term, $k$ verified normals\\[2pt]

\begin{adjustbox}{max width=\linewidth}
\begin{tabular}{@{}l|ccc|ccc@{}}
\toprule
& \multicolumn{3}{c|}{\textbf{AeBAD-S~\cite{AeBAD}}}
& \multicolumn{3}{c}{\textbf{RobustAD~\cite{RobustAD}}} \\
\midrule

Frozen detector ($k{=}0$) $\rightarrow$
& \multicolumn{3}{c|}{84.1/88.5}
& \multicolumn{3}{c}{83.0/90.3} \\
\midrule

Target term $\downarrow$
& $k{=}1$ & $k{=}2$ & $k{=}4$
& $k{=}1$ & $k{=}2$ & $k{=}4$ \\
\midrule

without the prior
& 89.3/90.2 & 89.2/90.1 & 89.4/90.2
& 87.1/93.2 & 88.6/\textbf{93.8} & 91.2/94.4 \\

convex mix with the prior
& 88.9/90.8 & 90.5/91.0 & 91.3/91.0
& 87.3/93.1 & 89.2/93.1 & 90.8/94.2 \\

mean only (no span)
& 90.1/91.2 & 90.7/\textbf{91.3} & 91.3/91.2
& \textbf{88.2}/93.2 & 89.0/93.7 & 90.4/94.1 \\

full weight ($w{=}1$)
& 88.5/90.1 & 91.3/90.9 & 91.5/\textbf{91.3}
& 86.5/90.6 & \textbf{89.3}/93.2 & 90.8/94.3 \\

\midrule
\rowcolor{oursbg}
\textbf{Ours (JS weight)}
& \textbf{90.2}/\textbf{91.3}
& \textbf{91.6/91.3}
& \textbf{91.7/91.3}
& \textbf{88.2/93.3}
& \textbf{89.3/93.8}
& \textbf{91.3/94.5} \\

\bottomrule
\end{tabular}
\end{adjustbox}
\end{minipage}

\end{table*}

\begin{table*}[t]
\centering
\scriptsize
\setlength{\tabcolsep}{3.2pt}
\renewcommand{\arraystretch}{1.12}
\setlength{\aboverulesep}{2pt}
\setlength{\belowrulesep}{1.5pt}
\caption{\textbf{Backbone ablation} with
$k$ verified target normals ($k=0$: source-only),
mean over domains and 5 draws.
The best in \textbf{bold}, the second-highest is
\underline{underlined}.
\textcolor{axisred}{Control} denotes the frozen detector.
Renormaly demonstrates broad robustness when
evaluated under matched backbones.
Resolution is matched per baseline for fair comparison.}
\label{tab:backbone}

\begin{tabular}{@{}ll*{12}{c}@{}}
\toprule
& & \multicolumn{6}{c}{\textbf{AeBAD-S~\cite{AeBAD}}}
  & \multicolumn{6}{c}{\textbf{RobustAD~\cite{RobustAD}}} \\
\cmidrule(lr){3-8}\cmidrule(lr){9-14}
& & \multicolumn{3}{c}{$k=0$}
  & \multicolumn{3}{c}{$k=1$}
  & \multicolumn{3}{c}{$k=0$}
  & \multicolumn{3}{c}{$k=1$} \\
\cmidrule(lr){3-5}\cmidrule(lr){6-8}
\cmidrule(lr){9-11}\cmidrule(lr){12-14}
Backbone & Method
& I-AUROC & P-AP & AUPRO
& I-AUROC & P-AP & AUPRO
& I-AUROC & P-AP & AUPRO
& I-AUROC & P-AP & AUPRO \\
\midrule

\multirow{4}{*}{WRN-50~\cite{WideResNet}}
& PatchCore~\cite{PatchCore}
& 71.1 & \underline{13.0} & 85.0
& 71.9 & \underline{13.0} & 85.1
& 70.8 & \textbf{8.0} & \textbf{71.1}
& 78.6 & 8.8 & 75.6 \\

& ADShift~\cite{ADShift}
& \underline{74.7} & 10.2 & \underline{89.0}
& \underline{78.3} & 10.4 & \underline{89.1}
& \underline{71.3} & \underline{6.8} & 67.3
& 81.9 & \underline{9.7} & \underline{84.9} \\

& FiCo~\cite{FiCO}
& 60.1 & 4.3 & 74.2
& 76.2 & 9.2 & 88.2
& 68.9 & 6.0 & 69.4
& \textbf{83.0} & 8.8 & \textbf{85.7} \\

& \textcolor{axisred}{Control}
& \textcolor{axisred}{72.2} & \textcolor{axisred}{8.6} & \textcolor{axisred}{87.1}
& \textcolor{axisred}{--} & \textcolor{axisred}{--} & \textcolor{axisred}{--}
& \textcolor{axisred}{70.4} & \textcolor{axisred}{4.2} & \textcolor{axisred}{66.9}
& \textcolor{axisred}{--} & \textcolor{axisred}{--} & \textcolor{axisred}{--} \\

\cmidrule(l){2-14}
\rowcolor{oursbg}
\cellcolor{white}
& \textbf{Renormaly (Ours)}
& \textbf{86.1} & \textbf{13.3} & \textbf{89.3}
& \textbf{86.4} & \textbf{13.7} & \textbf{89.4}
& \textbf{78.0} & \textbf{8.0} & \underline{70.2}
& \underline{82.7} & \textbf{9.8} & 77.3 \\

\midrule

\multirow{3}{*}{DINOv2-B-R~\cite{oquab2024dinov2}}
& INP-Former~\cite{INP-Former}
& 73.6 & 9.7 & 85.8
& 78.6 & 9.1 & 86.0
& 74.6 & \textbf{11.4} & 83.5
& \underline{81.8} & \textbf{12.7} & 86.6 \\

& Dinomaly~\cite{Dinomaly}
& \underline{79.5} & \underline{10.8} & \underline{86.8}
& \underline{83.0} & \underline{9.2} & \underline{86.5}
& 73.4 & 7.4 & 83.1
& 80.9 & 10.2 & \underline{88.7} \\

& \textcolor{axisred}{Control}
& \textcolor{axisred}{78.3} & \textcolor{axisred}{9.0} & \textcolor{axisred}{86.4}
& \textcolor{axisred}{--} & \textcolor{axisred}{--} & \textcolor{axisred}{--}
& \textcolor{axisred}{\underline{75.8}} & \textcolor{axisred}{8.8} & \textcolor{axisred}{\underline{83.6}}
& \textcolor{axisred}{--} & \textcolor{axisred}{--} & \textcolor{axisred}{--} \\

\cmidrule(l){2-14}
\rowcolor{oursbg}
\cellcolor{white}
& \textbf{Renormaly (Ours)}
& \textbf{90.6} & \textbf{11.0} & \textbf{91.9}
& \textbf{90.6} & \textbf{11.1} & \textbf{91.9}
& \textbf{85.4} & \underline{10.0} & \textbf{90.0}
& \textbf{87.0} & \underline{10.3} & \textbf{90.8} \\

\midrule

\multirow{2}{*}{DINOv3-B~\cite{DINOv3}}
& RAD~\cite{RAD2026}
& 74.5 & \textbf{12.8} & \underline{90.6}
& \underline{74.8} & \textbf{12.9} & \underline{90.7}
& 73.9 & \underline{12.2} & 81.8
& \underline{80.5} & \underline{14.1} & \underline{85.5} \\

& \textcolor{axisred}{Control}
& \textcolor{axisred}{\underline{84.1}} & \textcolor{axisred}{9.0} & \textcolor{axisred}{88.5}
& \textcolor{axisred}{--} & \textcolor{axisred}{--} & \textcolor{axisred}{--}
& \textcolor{axisred}{\underline{83.0}} & \textcolor{axisred}{10.2} & \textcolor{axisred}{\underline{90.3}}
& \textcolor{axisred}{--} & \textcolor{axisred}{--} & \textcolor{axisred}{--} \\

\cmidrule(l){2-14}
\rowcolor{oursbg}
\cellcolor{white}
& \textbf{Renormaly (Ours)}
& \textbf{87.6} & \underline{11.7} & \textbf{91.0}
& \textbf{90.2} & \underline{12.8} & \textbf{91.3}
& \textbf{85.4} & \textbf{16.9} & \textbf{92.8}
& \textbf{88.2} & \textbf{17.3} & \textbf{93.3} \\

\midrule

\multirow{2}{*}{AugReg~\cite{augreg}}

& \textcolor{axisred}{Control}
& \textcolor{axisred}{\underline{60.5}} & \textcolor{axisred}{\underline{4.4}} & \textcolor{axisred}{\underline{67.8}}
& \textcolor{axisred}{--} & \textcolor{axisred}{--} & \textcolor{axisred}{--}
& \textcolor{axisred}{\underline{46.5}} & \textcolor{axisred}{\underline{0.3}} & \textcolor{axisred}{\underline{63.1}}
& \textcolor{axisred}{--} & \textcolor{axisred}{--} & \textcolor{axisred}{--} \\

\cmidrule(l){2-14}
\rowcolor{oursbg}
\cellcolor{white}
& \textbf{Renormaly (Ours)}
& \textbf{71.4} & \textbf{5.8} & \textbf{74.2}
& \textbf{71.9} & \textbf{6.2} & \textbf{74.5}
& \textbf{54.3} & \textbf{1.8} & \textbf{68.5}
& \textbf{63.5} & \textbf{1.9} & \textbf{73.7} \\

\midrule

\multirow{2}{*}{DeiT~\cite{deit}}

& \textcolor{axisred}{Control}
& \textcolor{axisred}{\underline{62.4}} & \textcolor{axisred}{\underline{4.3}} & \textcolor{axisred}{\underline{67.5}}
& \textcolor{axisred}{--} & \textcolor{axisred}{--} & \textcolor{axisred}{--}
& \textcolor{axisred}{\underline{52.8}} & \textcolor{axisred}{\underline{0.1}} & \textcolor{axisred}{\underline{37.9}}
& \textcolor{axisred}{--} & \textcolor{axisred}{--} & \textcolor{axisred}{--} \\

\cmidrule(l){2-14}
\rowcolor{oursbg}
\cellcolor{white}
& \textbf{Renormaly (Ours)}
& \textbf{69.8} & \textbf{5.9} & \textbf{73.7}
& \textbf{70.2} & \textbf{6.2} & \textbf{74.0}
& \textbf{55.1} & \textbf{0.8} & \textbf{47.1}
& \textbf{61.9} & \textbf{1.0} & \textbf{52.1} \\

\midrule

\multirow{2}{*}{CLIP~\cite{clip}}

& \textcolor{axisred}{Control}
& \textcolor{axisred}{\underline{75.5}} & \textcolor{axisred}{\underline{5.6}} & \textcolor{axisred}{\underline{78.3}}
& \textcolor{axisred}{--} & \textcolor{axisred}{--} & \textcolor{axisred}{--}
& \textcolor{axisred}{\underline{54.9}} & \textcolor{axisred}{\underline{0.4}} & \textcolor{axisred}{\underline{60.1}}
& \textcolor{axisred}{--} & \textcolor{axisred}{--} & \textcolor{axisred}{--} \\

\cmidrule(l){2-14}
\rowcolor{oursbg}
\cellcolor{white}
& \textbf{Renormaly (Ours)}
& \textbf{82.2} & \textbf{6.3} & \textbf{81.6}
& \textbf{82.3} & \textbf{6.5} & \textbf{81.8}
& \textbf{57.1} & \textbf{1.1} & \textbf{67.7}
& \textbf{65.1} & \textbf{1.5} & \textbf{73.5} \\

\bottomrule
\end{tabular}
\end{table*}

\label{sec:abl}

We ablate \textit{Renormaly} in the order of its design choices, starting from the frozen detector and progressively introducing the source-only prior and target term (\cref{tab:abl}).

\textbf{Source-only prior.}
We first ask which directions the residual should act on. Random directions provide modest gains over the frozen detector, improving image AUROC by $1.2$ points on AeBAD-S and $0.9$ on RobustAD. PCA directions improve further, reaching $86.1$ and $85.0$, respectively. MIR with energy capping performs best, achieving \textcolor{axisblue}{$87.6/91.0$} on AeBAD-S and \textcolor{axisblue}{$85.4/92.8$} on RobustAD (image AUROC/AUPRO). This corresponds to image AUROC gains of \textcolor{axisred}{$3.5{\uparrow}$} and \textcolor{axisred}{$2.4{\uparrow}$} points over the frozen detector.

We next examine how much energy to remove along the selected directions. Fully projecting out $V$ reduces image AUROC by $1.1$ points on AeBAD-S and $1.8$ on RobustAD relative to energy capping. This is consistent with full projection also suppressing defect-relevant information along $V$, whereas capping attenuates only excess energy. We use $d=16$ directions per perturbation family, with sensitivity to the subspace rank reported in \cref{fig:ranksensitivity}.

\textbf{Target term.}
We next examine how to incorporate $k$ verified target normals into the source-only prior. Retaining the prior improves image AUROC over target-only fitting by up to \textcolor{axisred}{$2.4{\uparrow}$} and \textcolor{axisred}{$1.1{\uparrow}$} points on AeBAD-S and RobustAD, respectively, with the gap narrowing as more target normals become available. Our formulation further outperforms convex mixing by up to \textcolor{axisred}{$1.3{\uparrow}$} and \textcolor{axisred}{$0.9{\uparrow}$} points, while modeling the target span rather than only its mean adds up to \textcolor{axisred}{$0.9{\uparrow}$} points on both benchmarks. At $k{=}1$, James--Stein weighting improves over full weighting ($w{=}1$) by \textcolor{axisred}{$1.7{\uparrow}$} points on both datasets, supporting uncertainty-aware target correction.

Together, these design choices yield substantial improvements over the frozen detector. With just one verified target normal ($k{=}1$), image AUROC improves by \textcolor{axisred}{$6.1{\uparrow}$} and \textcolor{axisred}{$5.2{\uparrow}$} points on AeBAD-S and RobustAD, respectively. At $k=4$, the gains reach \textcolor{axisred}{$7.6{\uparrow}$} and \textcolor{axisred}{$8.3{\uparrow}$} points, alongside AUPRO improvements of \textcolor{axisred}{$2.8{\uparrow}$} and \textcolor{axisred}{$4.2{\uparrow}$} points. These results demonstrate the effectiveness of combining source-only correction with a small number of verified target normals.

\noindent\textbf{Backbone fairness and further experiments.}
For a fair comparison, we evaluate Renormaly using each baseline's corresponding backbone and input resolution, as shown in \cref{tab:backbone}.
Renormaly outperforms competing methods in most settings on both benchmarks, and transfers its robustness to various foundation backbones. These results suggest that the improvements are not merely attributable to stronger pretrained features, but also to more effective normal-reference normalization. Results on all metrics are provided in detail in the supplementary material.

\section{Conclusion}
Renormaly keeps a feature-based detector frozen and corrects only how its residuals are compared with normal: a source-only
prior caps the residual energy along estimated nuisance directions, and a few verified normals re-center the reference with an
empirical-Bayes weight estimated from the data. On real acquisition shifts it gives the best average detection and region-level
localization with zero to four target normals, and its gains generalize across various foundational teacher backbones. Robustness to acquisition shift
thus need not be learned into the representation: realigning the reference a detector compares against is a cheaper and more
controllable path to deployment.

\section{LLM Usage Statement}
We used Claude (Anthropic) to assist with manuscript writing and the refinement of our methodology, including its implementation. All motivations and ideas were developed by the authors, who take full responsibility for the final results and arguments presented in this paper.

{
    \small
    \bibliographystyle{ieeenat_fullname}
    \bibliography{main}

\begin{thebibliography}{38}
\providecommand{\natexlab}[1]{#1}
\providecommand{\url}[1]{\texttt{#1}}
\expandafter\ifx\csname urlstyle\endcsname\relax
  \providecommand{\doi}[1]{doi: #1}\else
  \providecommand{\doi}{doi: \begingroup \urlstyle{rm}\Url}\fi

\bibitem[Batzner et~al.(2024)Batzner, Heckler, and K{\"o}nig]{EfficientAD}
Kilian Batzner, Lars Heckler, and Rebecca K{\"o}nig.
\newblock {EfficientAD}: Accurate visual anomaly detection at millisecond-level
  latencies.
\newblock pages 128--138, 2024.

\bibitem[Belrose et~al.(2023)Belrose, Schneider-Joseph, Ravfogel, Cotterell,
  Raff, and Biderman]{LEACE}
Nora Belrose, David Schneider-Joseph, Shauli Ravfogel, Ryan Cotterell, Edward
  Raff, and Stella Biderman.
\newblock {LEACE}: Perfect linear concept erasure in closed form.
\newblock pages 66044--66063, 2023.

\bibitem[Bergmann et~al.(2019)Bergmann, Fauser, Sattlegger, and
  Steger]{MVTecAD}
Paul Bergmann, Michael Fauser, David Sattlegger, and Carsten Steger.
\newblock {MVTec AD}: A comprehensive real-world dataset for unsupervised
  anomaly detection.
\newblock In \emph{CVPR}, 2019.

\bibitem[Blackwell(1953)]{Blackwell1953}
David Blackwell.
\newblock Equivalent comparisons of experiments.
\newblock \emph{The Annals of Mathematical Statistics}, 24\penalty0
  (2):\penalty0 265--272, 1953.

\bibitem[Cao et~al.(2023)Cao, Zhu, and Pang]{ADShift}
Tri Cao, Jiawen Zhu, and Guansong Pang.
\newblock Anomaly detection under distribution shift.
\newblock In \emph{ICCV}, pages 6511--6523, 2023.

\bibitem[Chen et~al.(2020)Chen, Kornblith, Norouzi, and Hinton]{chen2020simclr}
Ting Chen, Simon Kornblith, Mohammad Norouzi, and Geoffrey Hinton.
\newblock A simple framework for contrastive learning of visual
  representations.
\newblock pages 1597--1607, 2020.

\bibitem[Chen et~al.(2025)Chen, Luo, Wang, Zhao, Su, and Men]{FiCO}
Zining Chen, Xingshuang Luo, Weiqiu Wang, Zhicheng Zhao, Fei Su, and Aidong
  Men.
\newblock Filter or compensate: Towards invariant representation from
  distribution shift for anomaly detection.
\newblock In \emph{AAAI}, 2025.

\bibitem[Cohen and Hoshen(2020)]{SPADE}
Niv Cohen and Yedid Hoshen.
\newblock Sub-image anomaly detection with deep pyramid correspondences.
\newblock \emph{arXiv preprint arXiv:2005.02357}, 2020.

\bibitem[Damm et~al.(2025)Damm, Laszkiewicz, Lederer, and Fischer]{AnomalyDINO}
Simon Damm, Mike Laszkiewicz, Johannes Lederer, and Asja Fischer.
\newblock {AnomalyDINO}: Boosting patch-based few-shot anomaly detection with
  {DINOv2}.
\newblock pages 1319--1329, 2025.

\bibitem[Defard et~al.(2021)Defard, Setkov, Loesch, and Audigier]{PaDiM}
Thomas Defard, Aleksandr Setkov, Angelique Loesch, and Romaric Audigier.
\newblock {PaDiM}: A patch distribution modeling framework for anomaly
  detection and localization.
\newblock In \emph{Pattern Recognition: ICPR International Workshops and
  Challenges}, pages 475--489, 2021.

\bibitem[Deng and Li(2022)]{RD4AD}
Hanqiu Deng and Xingyu Li.
\newblock Anomaly detection via reverse distillation from one-class embedding.
\newblock In \emph{CVPR}, pages 9737--9746, 2022.

\bibitem[Efron and Morris(1973)]{Efron01031973}
Bradley Efron and Carl Morris.
\newblock Stein's estimation rule and its competitors—an empirical bayes
  approach.
\newblock \emph{Journal of the American Statistical Association}, 68\penalty0
  (341):\penalty0 117--130, 1973.

\bibitem[Ganin et~al.(2016)Ganin, Ustinova, Ajakan, Germain, Larochelle,
  Laviolette, Marchand, and Lempitsky]{DANN}
Yaroslav Ganin, Evgeniya Ustinova, Hana Ajakan, Pascal Germain, Hugo
  Larochelle, Fran{\c{c}}ois Laviolette, Mario Marchand, and Victor Lempitsky.
\newblock Domain-adversarial training of neural networks.
\newblock \emph{Journal of Machine Learning Research}, 17\penalty0
  (59):\penalty0 1--35, 2016.

\bibitem[Guo et~al.(2025)Guo, Lu, Zhang, Chen, Li, and Liao]{Dinomaly}
Jia Guo, Shuai Lu, Weihang Zhang, Fang Chen, Huiqi Li, and Hongen Liao.
\newblock {Dinomaly}: The less is more philosophy in multi-class unsupervised
  anomaly detection.
\newblock In \emph{CVPR}, pages 20405--20415, 2025.

\bibitem[He et~al.(2022)He, Chen, Xie, Li, Doll\'ar, and Girshick]{MAE}
Kaiming He, Xinlei Chen, Saining Xie, Yanghao Li, Piotr Doll\'ar, and Ross
  Girshick.
\newblock Masked autoencoders are scalable vision learners.
\newblock In \emph{CVPR}, 2022.

\bibitem[Huang et~al.(2023)Huang, Goh, Gu, and Susskind]{MAST}
Chen Huang, Hanlin Goh, Jiatao Gu, and Josh Susskind.
\newblock {MAST}: Masked augmentation subspace training for generalizable
  self-supervised priors.
\newblock In \emph{ICLR}, 2023.

\bibitem[James et~al.(1961)James, Stein, et~al.]{james1961estimation}
William James, Charles Stein, et~al.
\newblock Estimation with quadratic loss.
\newblock In \emph{Proceedings of the fourth Berkeley symposium on mathematical
  statistics and probability}, pages 361--379, 1961.

\bibitem[Jeong et~al.(2023)Jeong, Zou, Kim, Zhang, Ravichandran, and
  Dabeer]{WinCLIP}
Jongheon Jeong, Yang Zou, Taewan Kim, Dongqing Zhang, Avinash Ravichandran, and
  Onkar Dabeer.
\newblock {WinCLIP}: Zero-/few-shot anomaly classification and segmentation.
\newblock In \emph{CVPR}, pages 19606--19616, 2023.

\bibitem[Jose et~al.(2025)Jose, Moutakanni, Kang, Baldassarre, Darcet, Xu, Li,
  Szafraniec, Ramamonjisoa, Oquab, Sim\'eoni, Vo, Labatut, and
  Bojanowski]{dinov2r}
Cijo Jose, Th\'eo Moutakanni, Dahyun Kang, Federico Baldassarre, Timoth\'ee
  Darcet, Hu Xu, Daniel Li, Marc Szafraniec, Micha\"el Ramamonjisoa, Maxime
  Oquab, Oriane Sim\'eoni, Huy~V. Vo, Patrick Labatut, and Piotr Bojanowski.
\newblock Dinov2 meets text: A unified framework for image- and pixel-level
  vision-language alignment.
\newblock In \emph{CVPR}, pages 24905--24916, 2025.

\bibitem[Kechris(1995)]{Kechris1995}
Alexander~S. Kechris.
\newblock \emph{Classical Descriptive Set Theory}.
\newblock Springer, New York, 1995.

\bibitem[Loshchilov and Hutter(2019)]{adamw}
Ilya Loshchilov and Frank Hutter.
\newblock Decoupled weight decay regularization.
\newblock In \emph{ICLR}, 2019.

\bibitem[Luo et~al.(2025)Luo, Cao, Yao, Zhang, Lou, Cheng, Shen, and
  Yu]{INP-Former}
Wei Luo, Yunkang Cao, Haiming Yao, Xiaotian Zhang, Jianan Lou, Yuqi Cheng,
  Weiming Shen, and Wenyong Yu.
\newblock Exploring intrinsic normal prototypes within a single image for
  universal anomaly detection.
\newblock In \emph{Proceedings of the IEEE/CVF Conference on Computer Vision
  and Pattern Recognition (CVPR)}, pages 9974--9983, 2025.

\bibitem[Oquab et~al.(2024)Oquab, Darcet, Moutakanni, Vo, Szafraniec, Khalidov,
  Fernandez, et~al.]{oquab2024dinov2}
Maxime Oquab, Timoth{\'e}e Darcet, Th{\'e}o Moutakanni, Huy~V. Vo, Marc
  Szafraniec, Vasil Khalidov, Pierre Fernandez, et~al.
\newblock {DINOv2}: Learning robust visual features without supervision.
\newblock 2024.

\bibitem[Pemula et~al.(2025)Pemula, Zhang, and Dabeer]{RobustAD}
Latha Pemula, Dongqing Zhang, and Onkar Dabeer.
\newblock Robust ad: A real world benchmark dataset for robustness in
  industrial anomaly detection.
\newblock In \emph{CVPRW}, pages 4086--4096, 2025.

\bibitem[Radford et~al.(2021)Radford, Kim, Hallacy, Ramesh, Goh, Agarwal,
  Sastry, Askell, Mishkin, Clark, Krueger, and Sutskever]{clip}
Alec Radford, Jong~Wook Kim, Chris Hallacy, Aditya Ramesh, Gabriel Goh,
  Sandhini Agarwal, Girish Sastry, Amanda Askell, Pamela Mishkin, Jack Clark,
  Gretchen Krueger, and Ilya Sutskever.
\newblock Learning transferable visual models from natural language
  supervision.
\newblock In \emph{Proceedings of the 38th International Conference on Machine
  Learning}, pages 8748--8763. PMLR, 2021.

\bibitem[Ravfogel et~al.(2020)Ravfogel, Elazar, Gonen, Twiton, and
  Goldberg]{NullItOut}
Shauli Ravfogel, Yanai Elazar, Hila Gonen, Michael Twiton, and Yoav Goldberg.
\newblock Null it out: Guarding protected attributes by iterative nullspace
  projection.
\newblock pages 7237--7256, 2020.

\bibitem[Roth et~al.(2022)Roth, Pemula, Zepeda, Sch{\"o}lkopf, Brox, and
  Gehler]{PatchCore}
Karsten Roth, Latha Pemula, Joaquin Zepeda, Bernhard Sch{\"o}lkopf, Thomas
  Brox, and Peter Gehler.
\newblock Towards total recall in industrial anomaly detection.
\newblock In \emph{CVPR}, pages 14318--14328, 2022.

\bibitem[Schneider et~al.(2020)Schneider, Rusak, Eck, Bringmann, Brendel, and
  Bethge]{Schneider2020}
Steffen Schneider, Evgenia Rusak, Luisa Eck, Oliver Bringmann, Wieland Brendel,
  and Matthias Bethge.
\newblock Improving robustness against common corruptions by covariate shift
  adaptation.
\newblock 2020.

\bibitem[Siméoni et~al.(2025)Siméoni, Vo, Seitzer, Baldassarre, Oquab, Jose,
  Khalidov, Szafraniec, Yi, Ramamonjisoa, Massa, Haziza, Wehrstedt, Wang,
  Darcet, Moutakanni, Sentana, Roberts, Vedaldi, Tolan, Brandt, Couprie,
  Mairal, Jégou, Labatut, and Bojanowski]{DINOv3}
Oriane Siméoni, Huy~V. Vo, Maximilian Seitzer, Federico Baldassarre, Maxime
  Oquab, Cijo Jose, Vasil Khalidov, Marc Szafraniec, Seungeun Yi, Michaël
  Ramamonjisoa, Francisco Massa, Daniel Haziza, Luca Wehrstedt, Jianyuan Wang,
  Timothée Darcet, Théo Moutakanni, Leonel Sentana, Claire Roberts, Andrea
  Vedaldi, Jamie Tolan, John Brandt, Camille Couprie, Julien Mairal, Hervé
  Jégou, Patrick Labatut, and Piotr Bojanowski.
\newblock Dinov3, 2025.

\bibitem[Stein(1981)]{Stein1981}
Charles~M. Stein.
\newblock Estimation of the mean of a multivariate normal distribution.
\newblock \emph{The Annals of Statistics}, 9\penalty0 (6):\penalty0 1135--1151,
  1981.

\bibitem[Steiner et~al.(2022)Steiner, Kolesnikov, Zhai, Wightman, Uszkoreit,
  and Beyer]{augreg}
Andreas~Peter Steiner, Alexander Kolesnikov, Xiaohua Zhai, Ross Wightman, Jakob
  Uszkoreit, and Lucas Beyer.
\newblock How to train your vit? data, augmentation, and regularization in
  vision transformers.
\newblock \emph{Transactions on Machine Learning Research}, 2022.

\bibitem[Touvron et~al.(2021)Touvron, Cord, Douze, Massa, Sablayrolles, and
  J{\'e}gou]{deit}
Hugo Touvron, Matthieu Cord, Matthijs Douze, Francisco Massa, Alexandre
  Sablayrolles, and Herv{\'e} J{\'e}gou.
\newblock Training data-efficient image transformers \& distillation through
  attention.
\newblock 2021.

\bibitem[Wang et~al.(2024)Wang, Zhu, Gao, Gan, Zhang, Gu, Qian, Chen, and
  Ma]{RealIAD}
Chengjie Wang, Wenbing Zhu, Bin-Bin Gao, Zhenye Gan, Jiangning Zhang, Zhihao
  Gu, Shuguang Qian, Mingang Chen, and Lizhuang Ma.
\newblock {Real-IAD}: A real-world multi-view dataset for benchmarking
  versatile industrial anomaly detection.
\newblock In \emph{CVPR}, pages 22883--22892, 2024.

\bibitem[You et~al.(2022)You, Cui, Shen, Yang, Lu, Zheng, and Le]{UniAD}
Zhiyuan You, Lei Cui, Yujun Shen, Kai Yang, Xin Lu, Yu Zheng, and Xinyi Le.
\newblock A unified model for multi-class anomaly detection.
\newblock pages 4571--4584, 2022.

\bibitem[Zagoruyko and Komodakis(2016)]{WideResNet}
Sergey Zagoruyko and Nikos Komodakis.
\newblock Wide residual networks.
\newblock In \emph{BMVC}, pages 87.1--87.12, 2016.

\bibitem[Zhang et~al.(2026)Zhang, Li, Henderson, Aragon-Camarasa, and
  Long]{RAD2026}
Xingwu Zhang, Guanxuan Li, Paul Henderson, Gerardo Aragon-Camarasa, and Zijun
  Long.
\newblock Is task-specific training necessary for anomaly detection?
\newblock 2026.

\bibitem[Zhang et~al.(2023)Zhang, Zhao, Zhang, Sun, and Chen]{AeBAD}
Zilong Zhang, Zhibin Zhao, Xingwu Zhang, Chuang Sun, and Xuefeng Chen.
\newblock Industrial anomaly detection with domain shift: A real-world dataset
  and masked multi-scale reconstruction.
\newblock \emph{Computers in Industry}, 151:\penalty0 103990, 2023.

\bibitem[Zou et~al.(2022)Zou, Jeong, Pemula, Zhang, and Dabeer]{Spot}
Yang Zou, Jongheon Jeong, Latha Pemula, Dongqing Zhang, and Onkar Dabeer.
\newblock {SPot-the-Difference} self-supervised pre-training for anomaly
  detection and segmentation.
\newblock In \emph{ECCV}, 2022.

\end{thebibliography}
}

\clearpage
\appendix
\renewcommand{\topfraction}{0.9}
\renewcommand{\dbltopfraction}{0.9}
\renewcommand{\bottomfraction}{0.6}
\renewcommand{\textfraction}{0.08}
\renewcommand{\floatpagefraction}{0.75}
\renewcommand{\dblfloatpagefraction}{0.75}
\setcounter{topnumber}{4}
\setcounter{dbltopnumber}{4}
\setcounter{totalnumber}{6}
\makeatletter
\setlength{\@fptop}{0pt}
\setlength{\@fpbot}{0pt plus 1fil}
\setlength{\@dblfptop}{0pt}
\setlength{\@dblfpbot}{0pt plus 1fil}
\makeatother
\maketitlesupplementary
\section*{Overview}

The supplementary material provides the following sections to further support and complement the main manuscript:

\begin{itemize}[label=---, leftmargin=2em, itemsep=0pt, topsep=2pt]
    \item \textbf{\cref{sec:supp_impl}} shows more details on the implementation.
    \item \textbf{\cref{sec:supp_proofs}} gives proofs of our propositions.
    \item \textbf{\cref{sec:runtime}} shows memory overhead and time complexity analysis of Renormaly.
    \item \textbf{\cref{sec:results}} shows extended results of the main manuscript and the influence of the seed and input resolution.
    \item \textbf{\cref{sec:limitation}} provides visualizations and analysis of failure cases of Renormaly. 
    
\end{itemize}

\section{Extended Implementation Details}
\label{sec:supp_impl}

\paragraph{Detector.}
The student is initialized from a pretrained MAE and has one light head per teacher block. During training it sees
the image with $40\%$ of its tokens masked and learns to predict the teacher features of all tokens with a cosine loss.
Tokens on the background count less in this loss: their weight is lowered by a foreground prior that the teacher itself
provides, from the similarity between each patch token and the \texttt{CLS} token in blocks $9$ and $11$. Each category is
trained for $200$ epochs with AdamW~\cite{adamw} (learning rate $10^{-3}$, batch size $16$). Images are resized to a square and
center-cropped to $224$ for AeBAD-S and MVTec-AD, $448$ for VisA and $512$ for RobustAD. The anomaly map is z-scored with
statistics of the source training images computed without calibration, so the calibration changes the map but not its
normalization.

\paragraph{Nuisance basis.}
The photometric interventions change brightness (two levels), contrast, gamma and color temperature, and add a horizontal or a
vertical shading ramp, seven in total. The five background interventions keep the foreground given by the prior and refill the
rest with color fills or with uniform noise. The basis is estimated once per category from the source
training images at $224$ pixels and reused at higher resolutions, since it acts on the feature channels and not on the token
grid.

\paragraph{Evaluation.}
\Cref{tab:supp_data} gives the test images of every domain. Each of the 5 draws sets aside, at random, 16 normal and 16
defective test images of every domain. The verified normals are the first $k$ of these 16 normals, so the sets for
$k{=}1,2,4$ are nested, and every method is scored on the remaining test images, the same set for every method and every $k$.
The defective pool is not used by any method. It was drawn because the split was designed for a broader protocol in which the
unlabeled target batch may also contain defects, and we kept the same split so that all our runs are paired on one scored set.
Excluding these images removes the same images for every method and only reduces the number of scored defects, so it cannot
favour any method.
For the pixel metrics, the anomaly map of every method is brought to the frame of the ground-truth masks and evaluated at
$224^2$ for AeBAD-S and at $256^2$ for RobustAD, where both the $512^2$ maps and masks are average-pooled by $2\times$ and the
masks re-binarised at $0.5$.

\begin{table}[t]\centering\small
\caption{\textbf{Number of test images per domain (normal / defective)}. }
\label{tab:supp_data}
\setlength{\tabcolsep}{3pt}
\begin{tabular}{@{}llrrrr@{}}
\toprule[1.5pt]
Benchmark & Domain & Test & Normal & Defective & Scored \\
\midrule
AeBAD-S    & same (source) & 689 & 230 & 459 & 657 \\
           & background    & 305 &  93 & 212 & 273 \\
           & illumination  & 273 &  75 & 198 & 241 \\
           & view          & 372 &  92 & 280 & 340 \\
\midrule
MetalParts & 0 (source)    & 171 & 120 &  51 & 139 \\
           & 1--4 (each)   & 200 & 120 &  80 & 168 \\
           & 5--6 (each)   & 100 &  60 &  40 &  68 \\
\midrule
PCB        & 0 (source)    & 195 & 120 &  75 & 163 \\
           & 1             & 210 & 120 &  90 & 178 \\
           & 2             & 180 & 120 &  60 & 148 \\
           & 3--5 (each)   & 140 &  80 &  60 & 108 \\
\midrule
PiledBags  & 0 (source)    & 195 & 120 &  75 & 163 \\
           & 1             & 210 & 120 &  90 & 178 \\
           & 2             & 420 & 240 & 180 & 388 \\
           & 3--5 (each)   & 210 & 120 &  90 & 178 \\
\midrule
\multicolumn{2}{@{}l}{Total (23 domains)} & 5{,}270 & 2{,}650 & 2{,}620 & 4{,}534 \\
\bottomrule[1.5pt]
\end{tabular}
\end{table}

\paragraph{Baselines.}
Every baseline runs its official code with its default single-class configuration: PatchCore~\cite{PatchCore} with WideResNet-50~\cite{WideResNet}, RAD~\cite{RAD2026}
with DINOv3~\cite{DINOv3} ViT-B/16 weights, Dinomaly~\cite{Dinomaly} and INP-Former~\cite{INP-Former} with DINOv2-R~\cite{dinov2r} ViT-B/14, and ADShift~\cite{ADShift} and FiCo~\cite{FiCO} with
WideResNet-50~\cite{WideResNet}. The two memory bank methods use the $k$ normals without training: PatchCore adds their patches to its
coreset, and RAD adds them to its memory, as in its own few-shot mode.

\paragraph{Fine-tuning the learned baselines.}
Dinomaly, INP-Former, ADShift and FiCo are fine-tuned on the $k$ normals with one recipe, fixed before any experiment and the
same for all four:
\begin{itemize}
\item \emph{Start and scope.} Fine-tuning starts from the source model and updates the same parts as source training. The
pre-trained encoder stays frozen.
\item \emph{Data.} The $k$ normals are the only target data. They are augmented with the few-shot augmentation of
INP-Former (rotations, translations, a flip and a grey-scale copy), so INP-Former is fine-tuned with its own
few-shot recipe. Dinomaly has no few-shot mode, and INP-Former is built on it with the same encoder, optimizer and scoring, so
the same recipe is used for Dinomaly and, for uniformity, for ADShift and FiCo, which keep their own AugMix and grey-scale
augmentation on top because their loss requires it. Ours uses the $k$ normals as they are.

\item \emph{Loss and optimizer.} Each method keeps its own, at a constant learning rate of one tenth of its base value.
\item \emph{No selection.} No target label is used and no checkpoint is chosen: the model after the last step is scored.
\end{itemize}
At $k=0$ every method is its source model.

\paragraph{Software and hardware.}
All methods run with Python 3.10 and PyTorch 2.6 on an NVIDIA RTX A6000 (48 GB) GPU.

\section{Proofs of the Theoretical Results}
\label{sec:supp_proofs}

\begin{table}[t]\centering\small
\setlength{\tabcolsep}{4pt}
\begin{tabular}{c|c}
\toprule[1.5pt]
Symbol & Meaning \\
\midrule
$z^T_\ell(x,p)$, $z^S_\ell(x,p)$ & teacher / student token in $\mathbb{R}^C$ \\
$r_\ell=z^T_\ell-z^S_\ell$, $\mu_\ell(p)$ & residual and its source training mean \\
$V_\ell=[v_{\ell1},\dots,v_{\ell,2d}]$ & orthonormal nuisance basis, Eq.~\eqref{eq:mirmom} \\
$c_\ell=V_\ell^\top(r_\ell-\mu_\ell)$ & residual coordinates along $V_\ell$ \\
$e_{\ell j}(x)$, $\lambda_{\ell j}$ & robust energy along $v_{\ell j}$, source level \\
$\Delta^0_\ell$, $\Delta^{\mathrm{tgt}}_\ell$, $\Delta_\ell$ & prior, target term, total correction \\
$m_\ell$, $U_\ell$, $\hat\Sigma_\ell$, $w_\ell$ & statistics of the $k$ normals, Eq.~\eqref{eq:eb} \\
$A_s$, $A_t$, $A_t^{\mathrm{ref}}$, $A^*_s$, $A^*_t$ & AUROCs of Sec.~\ref{sec:diagnosis} \\
\bottomrule[1.5pt]
\end{tabular}
\caption{Notation.}
\label{tab:supp_notation}
\end{table}

\begin{table*}[t]
\centering
\small
\setlength{\tabcolsep}{6pt}

\newcommand{\chg}[1]{\,{\scriptsize(\textcolor{axisred}{#1})}}

\begin{tabular}{@{}llcccccc@{}}
\toprule[1.5pt]
 & & \multicolumn{2}{c}{Inference time per image (ms)}
 & \multicolumn{2}{c}{Peak GPU memory (MiB)}
 & FLOPs & Stored \\
\cmidrule(lr){3-4}\cmidrule(lr){5-6}
Benchmark & Method & batch 1 & batch 16
 & batch 1 & batch 16 & (G) & (M values) \\
\midrule

\multirow{3}{*}{AeBAD-S ($\textcolor{axisblue}{224^2}$)}
 & Without Renormaly
 & 19.5 & 5.84 & 742 & 943 & 36.6 & 0 \\

 & Renormaly, $k{=}0$
 & 19.9\chg{+2.2\%}
 & 5.90\chg{+1.0\%}
 & 742\chg{+0.1\%}
 & 944\chg{+0.0\%}
 & 36.6\chg{+0.039} & 0.7 \\

 & Renormaly, $k{=}4$
 & 20.3\chg{+4.1\%}
 & 6.12\chg{+4.8\%}
 & 753\chg{+1.6\%}
 & 1029\chg{+9.1\%}
 & 36.6\chg{+0.042} & 3.1 \\

\midrule

\multirow{3}{*}{RobustAD ($\textcolor{axisblue}{512^2}$)}
 & Without Renormaly
 & 42.3 & 34.7 & 790 & 1916 & 189.0 & 0 \\

 & Renormaly, $k{=}0$
 & 42.7\chg{+1.0\%}
 & 35.0\chg{+0.9\%}
 & 791\chg{+0.1\%}
 & 1917\chg{+0.0\%}
 & 189.2\chg{+0.201} & 3.2 \\

 & Renormaly, $k{=}4$
 & 43.6\chg{+3.2\%}
 & 36.2\chg{+4.2\%}
 & 875\chg{+10.7\%}
 & 2360\chg{+23.2\%}
 & 189.2\chg{+0.220} & 15.8 \\

\bottomrule[1.5pt]
\end{tabular}

\caption{\textbf{Renormaly adds almost no cost.} It stays within $5\%$ of the detector's inference time and $0.12\%$ of its FLOPs
at both batch sizes, and needs no extra memory at $k{=}0$. Measured on one RTX A6000, in parentheses the change over the
detector without Renormaly.}
\label{tab:eff}
\end{table*}

\begin{table*}[t]
\centering
\footnotesize\setlength{\tabcolsep}{0pt}
\begin{tabular*}{\textwidth}{@{\extracolsep{\fill}}cl ccccccc ccccccc@{}}
\toprule[1.5pt]
 & & \multicolumn{7}{c}{AeBAD-S\cite{AeBAD}} & \multicolumn{7}{c}{RobustAD\cite{RobustAD}} \\
\cmidrule(lr){3-9}\cmidrule(lr){10-16}
Input & Method & I-AUROC & I-AP & I-F1 & P-AUROC & P-AP & P-F1 & AUPRO & I-AUROC & I-AP & I-F1 & P-AUROC & P-AP & P-F1 & AUPRO \\
\midrule
\multirow{3}{*}{$\textcolor{axisblue}{224^2}$} & Frozen detector & 84.1 & 93.3 & 90.3 & 93.3 & 9.0 & 15.8 & 88.5 & 78.8 & 74.8 & 76.2 & 92.4 & 2.3 & 11.4 & 72.5 \\
 & Renormaly, $k{=}0$ & 87.6 & 95.6 & 90.5 & 93.8 & 11.7 & 20.0 & 91.0 & 79.8 & 75.2 & 75.4 & 93.5 & 6.2 & 12.4 & 76.9 \\
 & Renormaly, $k{=}4$ & 91.7 & 96.7 & 91.7 & 94.1 & 13.2 & 20.3 & 91.3 & 88.4 & 84.3 & 81.9 & 95.0 & 6.7 & 13.1 & 81.8 \\
\midrule
\multirow{3}{*}{$\textcolor{axisblue}{320^2}$} & Frozen detector & 85.2 & 95.5 & 90.5 & 92.3 & 9.5 & 19.3 & 90.5 & 78.1 & 75.1 & 75.6 & 95.8 & 2.9 & 15.8 & 82.2 \\
 & Renormaly, $k{=}0$ & 87.4 & 95.5 & 90.9 & 92.7 & 12.0 & 20.1 & 92.2 & 80.1 & 77.0 & 74.9 & 96.2 & 9.6 & 17.4 & 85.3 \\
 & Renormaly, $k{=}4$ & 91.5 & 96.6 & 92.1 & 93.1 & 13.6 & 20.4 & 92.6 & 89.9 & 86.0 & 82.9 & 97.2 & 10.6 & 18.6 & 88.9 \\
\midrule
\multirow{3}{*}{$\textcolor{axisblue}{448^2}$} & Frozen detector & 84.6 & 95.1 & 91.6 & 91.3 & 9.4 & 18.7 & 90.5 & 82.7 & 79.4 & 76.7 & 98.3 & 9.0 & 22.4 & 90.0 \\
 & Renormaly, $k{=}0$ & 86.8 & 95.1 & 91.3 & 91.9 & 11.7 & 19.3 & 92.0 & 85.1 & 81.0 & 77.5 & 98.4 & 15.5 & 23.7 & 92.6 \\
 & Renormaly, $k{=}4$ & 90.9 & 96.1 & 92.5 & 92.2 & 13.3 & 19.7 & 92.4 & 91.3 & 87.1 & 84.0 & 98.8 & 16.7 & 24.8 & 94.3 \\
\midrule
\multirow{3}{*}{$\textcolor{axisblue}{512^2}$} & Frozen detector & 84.3 & 94.9 & 91.3 & 91.1 & 9.5 & 18.7 & 90.1 & 83.0 & 80.0 & 76.3 & 98.0 & 10.2 & 24.3 & 90.3 \\
 & Renormaly, $k{=}0$ & 86.8 & 95.1 & 91.5 & 91.6 & 11.9 & 19.6 & 91.7 & 85.4 & 80.7 & 77.9 & 98.4 & 16.9 & 25.8 & 92.8 \\
 & Renormaly, $k{=}4$ & 90.9 & 96.1 & 92.7 & 92.0 & 13.5 & 19.9 & 92.0 & 91.3 & 86.7 & 83.8 & 98.8 & 18.4 & 27.4 & 94.5 \\
\bottomrule[1.5pt]
\end{tabular*}
\caption{\textbf{Resolution sensitivity}. The student is retrained at each input resolution with every other
setting fixed. Results
reported in the main manuscript use $224^2$ for AeBAD-S and $512^2$ for RobustAD.}
\label{tab:resolution}
\end{table*}

\paragraph{Setting.}
Let $X$ be the frozen features of an image and $Y\in\{0,1\}$ its label (normal, defective). A \emph{test} flags the image
with probability $\varphi(X)\in[0,1]$; thresholding a score is a test. Its error rates are
$\mathrm{FPR}=\mathbb{E}[\varphi(X)\mid Y{=}0]$ and $\mathrm{TPR}=\mathbb{E}[\varphi(X)\mid Y{=}1]$. The \emph{best ROC curve}
of $X$ gives, at each FPR, the largest TPR of any test, and $A^*$ is the area under it. By the Neyman--Pearson lemma,
thresholding the likelihood ratio of $X$ attains this curve, so no score computed from $X$ has an AUROC above $A^*$. All AUROCs
in this section are population quantities.

\subsection{Loss Decomposition and Information-Loss Bound (Eqs.~\eqref{eq:split}--\eqref{eq:bound})}

\paragraph{Statement.} Eq.~\eqref{eq:split} holds exactly, and $A^*_s-A^*_t\le(A^*_s-A_s)+(1-\phi)(A_s-A_t)$.

\paragraph{Proof.} Eq.~\eqref{eq:split} is an identity: on its right-hand side, $A^*_s$, $A^*_t$ and $A_t^{\mathrm{ref}}$
each appear once with each sign, and what remains is $A_s-A_t$. The bound uses two facts.
\begin{enumerate}
\item By the definition of $\phi$, $A_s-A_t^{\mathrm{ref}}=(1-\phi)(A_s-A_t)$.
\item $A_t^{\mathrm{ref}}\le A^*_t$. The rebuilt reference uses normals drawn independently of the scored images, so the
detector scores each image by a fixed function of its features, and no such score exceeds $A^*_t$.
\end{enumerate}
Together,
\begin{equation*}
\begin{aligned}
A^*_s-A^*_t&\le A^*_s-A_t^{\mathrm{ref}}=(A^*_s-A_s)+(A_s-A_t^{\mathrm{ref}})\\
&=(A^*_s-A_s)+(1-\phi)(A_s-A_t).
\end{aligned}
\end{equation*}

\subsection{Information Limit under Shift (Proposition~\ref{prop:info})}

Let $X$ be the source features of an image and $X'$ its features in the shifted setup. The shift acts as $X'=g(X,N)$, where
the nuisance $N$ is independent of $(X,Y)$: it does not depend on the image or on whether it is defective.

\paragraph{Proposition.} The best ROC curve of $X'$ lies on or below that of $X$ at every FPR, so $A^*_t\le A^*_s$. The curves
coincide if the shift is the same invertible map $h$ for every image.

\paragraph{Proof.} \emph{Idea: whatever a test can do with $X'$, it can do with $X$ by simulating the nuisance.}
\begin{enumerate}
\item Take any test $\varphi'$ on $X'$. Define a test on $X$ by drawing a fresh nuisance and applying $\varphi'$:
$\varphi(X)=\mathbb{E}_N[\varphi'(g(X,N))]$.
\item Since $N$ is independent of $(X,Y)$, $\varphi(X)$ and $\varphi'(X')$ have the same average under each label:
\begin{equation*}
\mathbb{E}[\varphi(X)\mid Y=y]=\mathbb{E}[\varphi'(X')\mid Y=y],\quad y\in\{0,1\}.
\end{equation*}
So $\varphi$ has the same FPR and TPR as $\varphi'$.
\item Every (FPR, TPR) point reachable with $X'$ is thus reachable with $X$, and the best curve of $X'$ cannot be higher.
\item If $X'=h(X)$ for an invertible $h$, then $X=h^{-1}(X')$ ($h^{-1}$ is measurable~\cite{Kechris1995}), and the same
argument in reverse gives equality.
\end{enumerate}
This is Blackwell's comparison of experiments~\cite{Blackwell1953}. As it holds at every FPR, it also bounds the TPR at a fixed
FPR.

\paragraph{Proposition.} In the Gaussian model below, the best ROC curves before and after the shift coincide if and only if
$\Sigma_n\Sigma^{-1}a=0$, whatever the nuisance mean $m$.

\paragraph{Model.} Source features are $\mathcal{N}(\mu,\Sigma)$ for normal images and $\mathcal{N}(\mu+a,\Sigma)$ for a
defect of type $a$, with $\Sigma\succ0$. The shift adds an independent nuisance, $X'=X+n$ with $n\sim\mathcal{N}(m,\Sigma_n)$.
After the shift, normal images are $\mathcal{N}(\mu+m,\Sigma')$ and defective ones $\mathcal{N}(\mu+m+a,\Sigma')$, with
$\Sigma'=\Sigma+\Sigma_n$.

\paragraph{Proof.} \emph{Idea: in this model the best curve depends on a single number, the Mahalanobis separation.}
\begin{enumerate}
\item For two Gaussians with a common covariance $S$ and means $a$ apart, the best curve is
$\mathrm{TPR}=\Phi(\delta-\Phi^{-1}(1-\mathrm{FPR}))$ with $\delta^2=a^\top S^{-1}a$, and it rises with $\delta$. So the curves
coincide if and only if $\delta^2=a^\top\Sigma^{-1}a$ (source) equals $\delta'^2=a^\top\Sigma'^{-1}a$ (shifted). The mean $m$
moves both classes alike and does not enter.
\item The matrix $\Sigma^{-1}-\Sigma'^{-1}=\Sigma'^{-1}\Sigma_n\Sigma^{-1}$ is positive semi-definite, since $\Sigma'\succeq\Sigma$.
Hence
\begin{equation*}
\delta^2-\delta'^2=a^\top(\Sigma^{-1}-\Sigma'^{-1})\,a\ge0 .
\end{equation*}
\item For a positive semi-definite matrix $M$, $a^\top Ma=0$ if and only if $Ma=0$. So $\delta=\delta'$ if and only if
$\Sigma'^{-1}\Sigma_n\Sigma^{-1}a=0$, \ie, $\Sigma_n\Sigma^{-1}a=0$.
\end{enumerate}
Applied to each defect type, the curves coincide for all types in $\mathcal{A}$ if and only if the condition holds for every
$a\in\mathcal{A}$. The condition still suffices when the defective class mixes several types. For type $a$, the likelihood
ratio depends on the features only through $a^\top\Sigma^{-1}(X-\mu)$. The condition gives $\Sigma'^{-1}a=\Sigma^{-1}a$ and
leaves the nuisance no variance along $\Sigma^{-1}a$, so $a^\top\Sigma'^{-1}(X'-\mu-m)=a^\top\Sigma^{-1}(X-\mu)$ for every
image. The likelihood ratio of the mixture is then unchanged image by image, and so is its best curve.

\subsection{Minimal Energy Cap (Proposition~\ref{prop:cap})}

\paragraph{Proposition.} Among per-image rescalings of the coordinates by factors in $[0,1]$ that cap the robust energy at
$\lambda_{\ell j}$, $\Delta^0$ is the smallest at every token. It is zero when no energy exceeds its source level, and a defect
on fewer than half of the tokens cannot raise the energy above a quantile of the other tokens.

\paragraph{Setup.} Fix an image and a block, and drop $\ell$. A rescaling multiplies coordinate $c_j(p)$ by a factor
$s_j\in[0,1]$ at every token, which subtracts
\begin{equation*}
\Delta^s(p)=\textstyle\sum_j(1-s_j)\,c_j(p)\,v_j
\end{equation*}
from the residual. It \emph{caps} the energy if each rescaled coordinate has robust energy at most $\lambda_j$.

\paragraph{Proof.}
\begin{enumerate}
\item \emph{Decoupling.} $V$ has orthonormal columns, so $\|\Delta^s(p)\|^2=\sum_j(1-s_j)^2c_j(p)^2$. Each term
depends only on its own $s_j$, and so does its cap, so each direction can be treated separately.
\item \emph{Feasible factors.} The median of $|s_jc_j|$ is $s_j$ times the median of $|c_j|$, so the rescaled energy is
$s_j^2e_j$, and the cap holds if and only if $s_j\le\sqrt{\lambda_j/e_j}$.
\item \emph{Optimal factors.} $(1-s_j)^2c_j(p)^2$ decreases in $s_j$ at every token, so the best choice is the largest
allowed factor, $s^\star_j=\min\{1,\sqrt{\lambda_j/e_j}\}$. This is Eq.~\eqref{eq:prior}. If $e_j\le\lambda_j$, then
$s^\star_j=1$ and nothing is removed along $v_j$.
\item \emph{Breakdown of the median.} The median is the $\lceil P/2\rceil$-th smallest of the $P$ values $|c_j(p)|$. Let a defect
change these values arbitrarily on fewer than $P/2$ tokens. At least $\lceil P/2\rceil$ values are then unchanged, and adding
values can only lower an order statistic. So the median is at most the $\lceil P/2\rceil$-th smallest unchanged value, however
large the defect.
\end{enumerate}
Step 4 treats the defect as local. With global attention it can also perturb other tokens slightly.

\subsection{Risk of the Empirical-Bayes Weight (Proposition~\ref{prop:js})}

\paragraph{Proposition.} If the corrected residuals of $k\ge2$ normals are i.i.d.\ Gaussian with covariance $\Sigma$ in a block,
the weighted offset $w(m-\mu)$ has a lower expected squared error than $m-\mu$, for every target mean, whenever
\begin{equation*}
(\mathrm{tr}\,\Sigma)^2\ge4\lambda_{\max}(\Sigma)\,\mathrm{tr}\,\Sigma+2\,\mathrm{tr}(\Sigma^2)/(k-1).
\end{equation*}

\paragraph{Setup.} Fix a block and stack its $P$ tokens, so that a corrected residual is one vector of dimension $PC$, with
mean $\theta$ and covariance $\Sigma$. Eq.~\eqref{eq:eb} then involves three quantities:
\begin{itemize}
\item the offset $y=m-\mu\sim\mathcal{N}(\vartheta,\Sigma/k)$, whose true value is $\vartheta=\theta-\mu$;
\item its squared length $t=\|y\|^2=\sum_p\|m(p)-\mu(p)\|^2$, the denominator of $w$;
\item the estimated noise in it, $\tau=\mathrm{tr}\,\hat\Sigma/k=\sum_p\mathrm{tr}\,\hat\Sigma(p)/k$, the numerator.
\end{itemize}
So $w=(1-\tau/t)_+$, and we compare $\hat y=w\,y$ with $y$. The error of $y$ is $\mathbb{E}\|y-\vartheta\|^2=\mathrm{tr}\,\Sigma/k
=:\alpha$; we also write $\beta=\lambda_{\max}(\Sigma)/k$. For Gaussian samples, $\tau$ is independent of $y$, with
$\mathbb{E}\tau=\alpha$ and $\mathrm{Var}\,\tau=2\,\mathrm{tr}(\Sigma^2)/(k^2(k-1))$ (Wishart).

\paragraph{Proof.} \emph{Idea: Stein's identity writes the error of $\hat y$ as $\alpha$ plus a correction term, and the
condition makes that term negative.}
\begin{enumerate}
\item \emph{Stein's identity.} Write $\hat y=y-g(y)$, where $g(y)=\min(1,\tau/t)\,y$ is the part that the weight removes. For
fixed $\tau$, Stein's identity~\cite{Stein1981} gives
\begin{equation*}
\mathbb{E}\|\hat y-\vartheta\|^2=\alpha+\mathbb{E}\,U,\quad U=\|g\|^2-2\,\mathrm{tr}\big(\tfrac{\Sigma}{k}\nabla g\big).
\end{equation*}
It remains to show $\mathbb{E}\,U<0$.
\item \emph{Bound on $U$.} Computing $\nabla g$ and using $y^\top\Sigma y\le\lambda_{\max}(\Sigma)\,t$, with
$\psi(s)=s^2-2s(\alpha-2\beta)$:
\begin{equation*}
U\le\psi(\tau)/t\ \ \text{if } t>\tau,\qquad U=\psi(t)/t-4\beta\ \ \text{if } t\le\tau .
\end{equation*}
\item \emph{The condition.} With the moments of $\tau$,
\begin{equation*}
\mathbb{E}\,\psi(\tau)=\frac{1}{k^2}\Big[-(\mathrm{tr}\,\Sigma)^2+4\lambda_{\max}(\Sigma)\,\mathrm{tr}\,\Sigma
+\frac{2\,\mathrm{tr}(\Sigma^2)}{k-1}\Big],
\end{equation*}
which is $\le0$ exactly under the condition. The condition also gives $\alpha>2\beta$.
\item \emph{Expectation over $\tau$.} Fix $t$. If $t\le2(\alpha-2\beta)$, then $\psi\le0$ on $[0,t]$, and both cases of Step 2
give $U\le-4\beta\,\mathbf{1}\{\tau\ge t\}$. Otherwise $\psi$ increases beyond $t$, so $\psi(t)\le\psi(\tau)$ when $\tau\ge t$,
and $U\le\psi(\tau)/t-4\beta\,\mathbf{1}\{\tau\ge t\}$ in both cases; Step 3 then removes the first term on average. Either way
\begin{equation*}
\mathbb{E}\|\hat y-\vartheta\|^2\le\alpha-4\beta\,P(\tau\ge t)<\alpha,
\end{equation*}
since $\tau>0$ and $t$ is small with positive probability.
\end{enumerate}
Because $\mathrm{tr}(\Sigma^2)\le\lambda_{\max}(\Sigma)\,\mathrm{tr}\,\Sigma$, the condition holds whenever
$\mathrm{tr}\,\Sigma/\lambda_{\max}(\Sigma)\ge4+2/(k-1)$. The proposition bounds the squared error of the offset, not AUROC,
and it does not cover the span term of Eq.~\eqref{eq:tgt}, which the same weight scales.

\section{Memory Overhead and Time Complexity}
\label{sec:runtime}

Renormaly adds no trainable parameters. With $|\mathcal{B}|$ teacher blocks, $P$ tokens of dimension $C$, $2d$ nuisance directions
per block and $k$ verified normals, it costs \textcolor{axisblue}{$O(|\mathcal{B}|PC(2d+k))$} per image on top of the \textcolor{axisblue}{$O(P^2C+PC^2)$} transformer
forward pass, \ie at most $0.22$\,G FLOPs ($0.12\%$) in \cref{tab:eff}. Besides the weights, it stores only $V$ and $\mu$
($0.7$\,M values on AeBAD-S, $3.2$\,M on RobustAD), plus the statistics of the $k$ normals for $k{>}0$.
We measure on one RTX A6000 at two batch sizes: batch 1 is the latency of inspecting one part at a time on a production line,
and batch 16 is the throughput setting, where fixed per-call costs such as kernel launches are shared across the batch and the
time per image mainly reflects the computation itself.
Renormaly adds at most $2.2\%$ inference time at $k{=}0$ and $4.8\%$ at $k{=}4$ in either setting.
The energy cap leaves peak memory unchanged. Re-centering with $k{=}4$ adds memory that grows with the batch size, from $1.6\%$ and $10.7\%$ at batch 1
to $9\%$ on AeBAD-S and $23\%$ on RobustAD at batch 16 ($444$\,MiB, total $2.4$\,GB).

\begin{table*}[t]
\centering
\scriptsize
\setlength{\tabcolsep}{3.0pt}
\renewcommand{\arraystretch}{1.08}
\setlength{\aboverulesep}{3pt}
\setlength{\belowrulesep}{1.5pt}
\caption{\textbf{Backbone-wise comparison of anomaly detection and localization performance on AeBAD-S~\cite{AeBAD} and RobustAD~\cite{RobustAD} under distribution shift.} Results are shown for source-only ($k{=}0$) and one verified target normal ($k{=}1$). AUPRO is computed up to FPR 0.3. Within each backbone and $k$ setting, the highest and second-highest distinct values per metric are highlighted in \textbf{bold} and \underline{underline}, respectively. Frozen detector results are shown in \textcolor{axisred}{red}.}
\label{tab:backbone_detailed}
\begin{tabular}{@{}cll*{14}{c}@{}}
\toprule
 & & & \multicolumn{3}{c}{\textbf{AeBAD-S \textit{(image-level)}}} & \multicolumn{4}{c}{\textbf{AeBAD-S \textit{(pixel-level)}}} & \multicolumn{3}{c}{\textbf{RobustAD \textit{(image-level)}}} & \multicolumn{4}{c}{\textbf{RobustAD \textit{(pixel-level)}}} \\
\cmidrule(lr){4-6}\cmidrule(lr){7-10}\cmidrule(lr){11-13}\cmidrule(lr){14-17}
\textbf{\# $k$} & \textbf{Backbone} & \textbf{Method} & AUROC & AP & F1 & AUROC & AP & F1 & AUPRO & AUROC & AP & F1 & AUROC & AP & F1 & AUPRO \\
\midrule
\cellcolor{white} & \cellcolor{white} & PatchCore~\cite{PatchCore} & 71.1 & 86.5 & 85.9 & \textbf{93.4} & \underline{13.0} & \textbf{20.2} & 85.0 & 70.8 & 67.1 & 68.8 & \underline{91.1} & \textbf{8.0} & \textbf{15.4} & \textbf{71.1} \\
\cellcolor{white} & \cellcolor{white} & ADShift~\cite{ADShift} & \underline{74.7} & 88.9 & 86.3 & 91.1 & 10.2 & 15.2 & \underline{89.0} & \underline{71.3} & 69.7 & 72.1 & 89.5 & \underline{6.8} & 12.0 & 67.3 \\
\cellcolor{white} & \cellcolor{white} & FiCo~\cite{FiCO} & 60.1 & 79.8 & 84.6 & 88.8 & 4.3 & 9.1 & 74.2 & 68.9 & 66.7 & 71.0 & 90.4 & 6.0 & 11.7 & 69.4 \\
\cellcolor{white} & \cellcolor{white} & \textcolor{axisred}{Frozen detector} & \textcolor{axisred}{72.2} & \textcolor{axisred}{\underline{92.4}} & \textcolor{axisred}{\underline{89.0}} & \textcolor{axisred}{90.3} & \textcolor{axisred}{8.6} & \textcolor{axisred}{14.9} & \textcolor{axisred}{87.1} & \textcolor{axisred}{70.4} & \textcolor{axisred}{\underline{75.5}} & \textcolor{axisred}{\underline{77.8}} & \textcolor{axisred}{90.2} & \textcolor{axisred}{4.2} & \textcolor{axisred}{6.8} & \textcolor{axisred}{66.9} \\
\cmidrule(l){3-17}
\rowcolor{oursbg}
\cellcolor{white} & \cellcolor{white}\multirow{-5}{*}{WRN-50} & \textbf{Renormaly (Ours)} & \textbf{86.1} & \textbf{94.6} & \textbf{89.9} & \underline{92.9} & \textbf{13.3} & \underline{15.9} & \textbf{89.3} & \textbf{78.0} & \textbf{77.1} & \textbf{78.2} & \textbf{91.6} & \textbf{8.0} & \underline{15.2} & \underline{70.2} \\
\cmidrule(l){2-17}
\cellcolor{white} & \cellcolor{white} & INP-Former~\cite{INP-Former} & 73.6 & 87.0 & 88.4 & 90.4 & 9.7 & 16.4 & 85.8 & 74.6 & 71.3 & 74.7 & \underline{95.5} & \textbf{11.4} & \underline{21.4} & 83.5 \\
\cellcolor{white} & \cellcolor{white} & Dinomaly~\cite{Dinomaly} & \underline{79.5} & \underline{90.1} & 89.7 & 90.9 & \underline{10.8} & \underline{17.6} & \underline{86.8} & 73.4 & 68.2 & 74.3 & 95.1 & 7.4 & 15.8 & 83.1 \\
\cellcolor{white} & \cellcolor{white} & \textcolor{axisred}{Frozen detector} & \textcolor{axisred}{78.3} & \textcolor{axisred}{89.9} & \textcolor{axisred}{\underline{89.8}} & \textcolor{axisred}{\underline{93.5}} & \textcolor{axisred}{9.0} & \textcolor{axisred}{17.0} & \textcolor{axisred}{86.4} & \textcolor{axisred}{\underline{75.8}} & \textcolor{axisred}{\underline{81.8}} & \textcolor{axisred}{\underline{80.3}} & \textcolor{axisred}{95.3} & \textcolor{axisred}{8.8} & \textcolor{axisred}{17.4} & \textcolor{axisred}{\underline{83.6}} \\
\cmidrule(l){3-17}
\rowcolor{oursbg}
\cellcolor{white} & \cellcolor{white}\multirow{-4}{*}{DINOv2-B-R} & \textbf{Renormaly (Ours)} & \textbf{90.6} & \textbf{95.9} & \textbf{91.7} & \textbf{94.5} & \textbf{11.0} & \textbf{18.7} & \textbf{91.9} & \textbf{85.4} & \textbf{84.1} & \textbf{81.9} & \textbf{97.1} & \underline{10.0} & \textbf{21.6} & \textbf{90.0} \\
\cmidrule(l){2-17}
\cellcolor{white} & \cellcolor{white} & RAD~\cite{RAD2026} & 74.5 & 88.0 & 87.1 & 92.6 & \textbf{12.8} & \underline{16.9} & \underline{90.6} & 73.9 & 70.8 & 74.0 & 95.1 & \underline{12.2} & 20.2 & 81.8 \\
\cellcolor{white} & \cellcolor{white} & \textcolor{axisred}{Frozen detector} & \textcolor{axisred}{\underline{84.1}} & \textcolor{axisred}{\underline{93.3}} & \textcolor{axisred}{\underline{90.3}} & \textcolor{axisred}{\underline{93.4}} & \textcolor{axisred}{9.0} & \textcolor{axisred}{15.8} & \textcolor{axisred}{88.5} & \textcolor{axisred}{\underline{83.0}} & \textcolor{axisred}{\underline{80.0}} & \textcolor{axisred}{\underline{76.3}} & \textcolor{axisred}{\underline{98.0}} & \textcolor{axisred}{10.2} & \textcolor{axisred}{\underline{24.3}} & \textcolor{axisred}{\underline{90.3}} \\
\cmidrule(l){3-17}
\rowcolor{oursbg}
\cellcolor{white} & \cellcolor{white}\multirow{-3}{*}{DINOv3-B} & \textbf{Renormaly (Ours)} & \textbf{87.6} & \textbf{95.6} & \textbf{90.5} & \textbf{93.8} & \underline{11.7} & \textbf{20.0} & \textbf{91.0} & \textbf{85.4} & \textbf{80.7} & \textbf{77.9} & \textbf{98.4} & \textbf{16.9} & \textbf{25.8} & \textbf{92.8} \\
\cmidrule(l){2-17}
\cellcolor{white} & \cellcolor{white} & \textcolor{axisred}{Frozen detector} & \textcolor{axisred}{\underline{60.5}} & \textcolor{axisred}{\underline{77.8}} & \textcolor{axisred}{\underline{85.1}} & \textcolor{axisred}{\underline{82.3}} & \textcolor{axisred}{\underline{4.4}} & \textcolor{axisred}{\underline{9.5}} & \textcolor{axisred}{\underline{67.8}} & \textcolor{axisred}{\underline{46.5}} & \textcolor{axisred}{\underline{41.1}} & \textcolor{axisred}{\underline{56.2}} & \textcolor{axisred}{\underline{87.3}} & \textcolor{axisred}{\underline{0.3}} & \textcolor{axisred}{\underline{0.9}} & \textcolor{axisred}{\underline{63.1}} \\
\cmidrule(l){3-17}
\rowcolor{oursbg}
\cellcolor{white} & \cellcolor{white}\multirow{-2}{*}{AugReg} & \textbf{Renormaly (Ours)} & \textbf{71.4} & \textbf{84.3} & \textbf{85.8} & \textbf{86.5} & \textbf{5.8} & \textbf{11.8} & \textbf{74.2} & \textbf{54.3} & \textbf{48.3} & \textbf{58.1} & \textbf{89.6} & \textbf{1.8} & \textbf{3.4} & \textbf{68.5} \\
\cmidrule(l){2-17}
\cellcolor{white} & \cellcolor{white} & \textcolor{axisred}{Frozen detector} & \textcolor{axisred}{\underline{62.4}} & \textcolor{axisred}{\underline{81.2}} & \textcolor{axisred}{\underline{84.9}} & \textcolor{axisred}{\underline{83.4}} & \textcolor{axisred}{\underline{4.3}} & \textcolor{axisred}{\underline{9.2}} & \textcolor{axisred}{\underline{67.5}} & \textcolor{axisred}{\underline{52.8}} & \textcolor{axisred}{\underline{46.5}} & \textcolor{axisred}{\textbf{60.3}} & \textcolor{axisred}{\underline{71.2}} & \textcolor{axisred}{\underline{0.1}} & \textcolor{axisred}{\underline{0.5}} & \textcolor{axisred}{\underline{37.9}} \\
\cmidrule(l){3-17}
\rowcolor{oursbg}
\cellcolor{white} & \cellcolor{white}\multirow{-2}{*}{DeiT} & \textbf{Renormaly (Ours)} & \textbf{69.8} & \textbf{84.3} & \textbf{86.1} & \textbf{86.8} & \textbf{5.9} & \textbf{11.4} & \textbf{73.7} & \textbf{55.1} & \textbf{48.1} & \textbf{60.3} & \textbf{77.3} & \textbf{0.8} & \textbf{3.0} & \textbf{47.1} \\
\cmidrule(l){2-17}
\cellcolor{white} & \cellcolor{white} & \textcolor{axisred}{Frozen detector} & \textcolor{axisred}{\underline{75.5}} & \textcolor{axisred}{\underline{89.3}} & \textcolor{axisred}{\underline{86.7}} & \textcolor{axisred}{\underline{88.1}} & \textcolor{axisred}{\underline{5.6}} & \textcolor{axisred}{\underline{11.3}} & \textcolor{axisred}{\underline{78.3}} & \textcolor{axisred}{\underline{54.9}} & \textcolor{axisred}{\underline{50.8}} & \textcolor{axisred}{\underline{62.0}} & \textcolor{axisred}{\underline{84.5}} & \textcolor{axisred}{\underline{0.4}} & \textcolor{axisred}{\underline{1.8}} & \textcolor{axisred}{\underline{60.1}} \\
\cmidrule(l){3-17}
\rowcolor{oursbg}
\cellcolor{white}\multirow{-18}{*}{\rotatebox{90}{$k{=}0$}} & \cellcolor{white}\multirow{-2}{*}{CLIP} & \textbf{Renormaly (Ours)} & \textbf{82.2} & \textbf{92.7} & \textbf{88.2} & \textbf{89.9} & \textbf{6.3} & \textbf{12.0} & \textbf{81.6} & \textbf{57.1} & \textbf{52.5} & \textbf{63.8} & \textbf{88.6} & \textbf{1.1} & \textbf{4.3} & \textbf{67.8} \\
\midrule
\cellcolor{white} & \cellcolor{white} & PatchCore~\cite{PatchCore} & 71.9 & 86.8 & 86.1 & \textbf{93.6} & \underline{13.0} & \textbf{20.2} & 85.1 & 78.6 & 73.7 & 73.3 & 92.9 & 8.8 & 16.4 & 75.6 \\
\cellcolor{white} & \cellcolor{white} & ADShift~\cite{ADShift} & \underline{78.3} & \underline{90.5} & 87.7 & 91.3 & 10.4 & 15.0 & \underline{89.1} & 81.9 & 76.9 & 77.1 & \underline{96.0} & \underline{9.7} & \textbf{19.1} & \underline{84.9} \\
\cellcolor{white} & \cellcolor{white} & FiCo~\cite{FiCO} & 76.2 & 87.8 & \underline{88.2} & 91.5 & 9.2 & 13.8 & 88.2 & \textbf{83.0} & \underline{78.3} & \underline{77.7} & \textbf{96.2} & 8.8 & \underline{18.2} & \textbf{85.7} \\
\cmidrule(l){3-17}
\rowcolor{oursbg}
\cellcolor{white} & \cellcolor{white}\multirow{-4}{*}{WRN-50} & \textbf{Renormaly (Ours)} & \textbf{86.4} & \textbf{94.7} & \textbf{90.1} & \underline{93.2} & \textbf{13.7} & \underline{15.9} & \textbf{89.4} & \underline{82.7} & \textbf{81.0} & \textbf{80.5} & 93.3 & \textbf{9.8} & 15.9 & 77.3 \\
\cmidrule(l){2-17}
\cellcolor{white} & \cellcolor{white} & INP-Former~\cite{INP-Former} & 78.6 & 89.8 & 89.2 & \underline{90.4} & 9.1 & 15.2 & 86.0 & \underline{81.8} & \underline{74.9} & 76.8 & 96.3 & \textbf{12.7} & \textbf{22.9} & 86.6 \\
\cellcolor{white} & \cellcolor{white} & Dinomaly~\cite{Dinomaly} & \underline{83.0} & \underline{92.2} & \underline{89.5} & \underline{90.4} & \underline{9.2} & \underline{15.5} & \underline{86.5} & 80.9 & 72.4 & \underline{77.6} & \underline{96.9} & 10.2 & 19.5 & \underline{88.7} \\
\cmidrule(l){3-17}
\rowcolor{oursbg}
\cellcolor{white} & \cellcolor{white}\multirow{-3}{*}{DINOv2-B-R} & \textbf{Renormaly (Ours)} & \textbf{90.6} & \textbf{95.9} & \textbf{91.8} & \textbf{94.5} & \textbf{11.1} & \textbf{18.8} & \textbf{91.9} & \textbf{87.0} & \textbf{85.2} & \textbf{83.3} & \textbf{97.4} & \underline{10.3} & \underline{22.5} & \textbf{90.8} \\
\cmidrule(l){2-17}
\cellcolor{white} & \cellcolor{white} & RAD~\cite{RAD2026} & \underline{74.8} & \underline{88.2} & \underline{87.2} & \underline{92.7} & \textbf{12.9} & \underline{16.9} & \underline{90.7} & \underline{80.5} & \underline{74.7} & \underline{76.5} & \underline{96.3} & \underline{14.1} & \underline{22.1} & \underline{85.5} \\
\cmidrule(l){3-17}
\rowcolor{oursbg}
\cellcolor{white} & \cellcolor{white}\multirow{-2}{*}{DINOv3-B} & \textbf{Renormaly (Ours)} & \textbf{90.2} & \textbf{96.7} & \textbf{91.5} & \textbf{94.1} & \underline{12.8} & \textbf{20.1} & \textbf{91.3} & \textbf{88.2} & \textbf{83.3} & \textbf{80.0} & \textbf{98.5} & \textbf{17.3} & \textbf{26.2} & \textbf{93.3} \\
\cmidrule(l){2-17}
\rowcolor{oursbg}
\cellcolor{white} & \cellcolor{white} AugReg & \textbf{Renormaly (Ours)} & 71.9 & 84.5 & 86.1 & 87.4 & 6.2 & 11.7 & 74.5 & 63.5 & 54.7 & 62.6 & 91.4 & 1.9 & 4.5 & 73.7 \\
\cmidrule(l){2-17}
\rowcolor{oursbg}
\cellcolor{white} & \cellcolor{white} DeiT & \textbf{Renormaly (Ours)} & 70.2 & 84.4 & 86.3 & 87.1 & 6.2 & 11.3 & 74.0 & 61.9 & 53.3 & 62.5 & 80.2 & 1.0 & 3.8 & 52.1 \\
\cmidrule(l){2-17}
\rowcolor{oursbg}
\cellcolor{white}\multirow{-12}{*}{\rotatebox{90}{$k{=}1$}} & \cellcolor{white} CLIP & \textbf{Renormaly (Ours)} & 82.3 & 92.7 & 88.3 & 90.1 & 6.5 & 12.0 & 81.8 & 65.1 & 57.6 & 67.2 & 90.9 & 1.5 & 5.9 & 73.5 \\
\bottomrule
\end{tabular}
\end{table*}

\section{Extended Results}
\label{sec:results}

\begin{figure*}[t]
  \centering
  \includegraphics[width=0.85\textwidth]{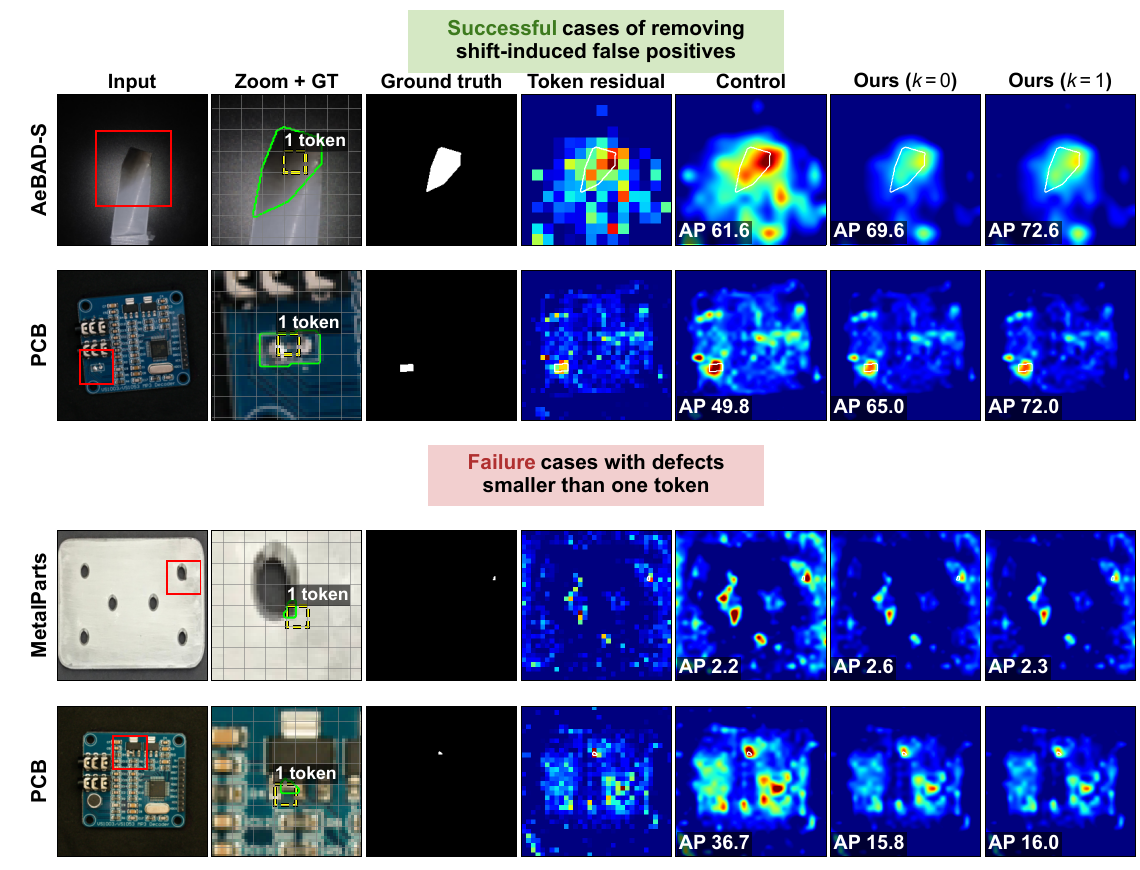}
  \caption{\textbf{Successful and failure cases} of Renormaly ($k{=}0$). The calibration removes shift-induced false
positives, so the defect stands out over the control (top). \textbf{Defects smaller than one ViT token remain hard (bottom)}: the
token that contains the defect responds, but most of its pixels are normal, which caps the pixel AP regardless of
calibration.}
\label{fig:supp_cases}
\label{fig:failures}

\end{figure*}

\subsection{Influence of Seed Variance}
\Cref{tab:seeds} reports the mean and standard deviation over five random seeds. Renormaly exhibits low run-to-run variance across both datasets, with standard deviations below 0.4 for all metrics. The consistent improvements over the frozen detector suggest that the gains from additional recalibration are robust across random seeds.

\begin{table*}[t]
\centering
\scriptsize\setlength{\tabcolsep}{0pt}
\begin{tabular*}{\textwidth}{@{\extracolsep{\fill}}l ccccccc ccccccc@{}}
\toprule[1.5pt]
 & \multicolumn{7}{c}{AeBAD-S\cite{AeBAD}} & \multicolumn{7}{c}{RobustAD\cite{RobustAD}} \\
\cmidrule(lr){2-8}\cmidrule(lr){9-15}
Method & I-AUROC & I-AP & I-F1 & P-AUROC & P-AP & P-F1 & AUPRO & I-AUROC & I-AP & I-F1 & P-AUROC & P-AP & P-F1 & AUPRO \\
\midrule
Frozen detector
& 84.1$\pm$\textcolor{axisblue}{0.3} & 93.3$\pm$\textcolor{axisblue}{0.1} & 90.3 $\pm$\textcolor{axisblue}{0.2} & 93.3$\pm$\textcolor{axisblue}{0.1} & 9.0$\pm$\textcolor{axisblue}{0.1} & 15.8$\pm$\textcolor{axisblue}{0.2} & 88.5$\pm$\textcolor{axisblue}{0.1}
& 83.0$\pm$\textcolor{axisblue}{0.4} & 80.0$\pm$\textcolor{axisblue}{0.4} & 76.3$\pm$\textcolor{axisblue}{0.4} & 98.0$\pm$\textcolor{axisblue}{0.0} & 10.2$\pm$\textcolor{axisblue}{0.1} & 24.3 $\pm$\textcolor{axisblue}{0.2} & 90.3$\pm$\textcolor{axisblue}{0.1} \\
Renormaly, $k{=}0$
& 87.6$\pm$\textcolor{axisblue}{0.3} & 95.6$\pm$\textcolor{axisblue}{0.1} & 90.5$\pm$\textcolor{axisblue}{0.2} & 93.8$\pm$\textcolor{axisblue}{0.1} & 11.7$\pm$\textcolor{axisblue}{0.1} & 20.0$\pm$\textcolor{axisblue}{0.2} & 91.0$\pm$\textcolor{axisblue}{0.1}
& 85.4$\pm$\textcolor{axisblue}{0.3} & 80.7$\pm$\textcolor{axisblue}{0.2} & 77.9$\pm$\textcolor{axisblue}{0.3} & 98.4$\pm$\textcolor{axisblue}{0.0} & 16.9$\pm$\textcolor{axisblue}{0.1} & 25.8$\pm$\textcolor{axisblue}{0.2} & 92.8$\pm$\textcolor{axisblue}{0.1} \\
Renormaly, $k{=}4$
& 91.7$\pm$\textcolor{axisblue}{0.3} & 96.7$\pm$\textcolor{axisblue}{0.1} & 91.7$\pm$\textcolor{axisblue}{0.2} & 94.1$\pm$\textcolor{axisblue}{0.1} & 13.2$\pm$\textcolor{axisblue}{0.1} & 20.3$\pm$\textcolor{axisblue}{0.2} & 91.3$\pm$\textcolor{axisblue}{0.1}
& 91.3$\pm$\textcolor{axisblue}{0.1} & 86.7$\pm$\textcolor{axisblue}{0.1} & 83.8$\pm$\textcolor{axisblue}{0.2} & 98.8$\pm$\textcolor{axisblue}{0.0} & 18.4$\pm$\textcolor{axisblue}{0.1} & 27.4$\pm$\textcolor{axisblue}{0.2} & 94.5$\pm$\textcolor{axisblue}{0.0} \\
\bottomrule[1.5pt]
\end{tabular*}
\caption{\textbf{Seed variance}. Mean\,$\pm$\,standard deviation over five seeds. Renormaly exhibits low run-to-run variance across both datasets, while increasing $k$ from 0 to 4 consistently improves performance.}
\label{tab:seeds}
\end{table*}

\subsection{Resolution Sensitivity}
\Cref{tab:resolution} retrains the student at each input resolution with all other settings fixed. With $k{=}4$, Renormaly improves over
the frozen detector on every metric at every resolution; RobustAD localization benefits most from higher resolution.

\begin{table*}[t]
\centering
\caption{\textbf{Per-domain (AeBAD-S) and per-category (RobustAD) results of Renormaly} (Extended of \cref{tab:main}).}
\label{tab:percat}

\begin{minipage}[t]{0.49\textwidth}
\centering
\scriptsize
\setlength{\tabcolsep}{2.4pt}
\renewcommand{\arraystretch}{1.05}
\setlength{\aboverulesep}{1.2pt}
\setlength{\belowrulesep}{1.2pt}

\begin{tabular}{@{}lc ccc cccc@{}}
\toprule
\multicolumn{9}{c}{\textbf{AeBAD-S}~\cite{AeBAD}} \\
\midrule
 & & \multicolumn{3}{c}{Image-level}
     & \multicolumn{4}{c}{Pixel-level} \\
\cmidrule(lr){3-5}\cmidrule(l){6-9}
Domain & $k$ & AUROC & AP & F1
       & AUROC & AP & F1 & AUPRO \\
\midrule

 & 0 & 85.2 & 93.3 & 87.2 & 94.6 & 10.0 & 18.0 & 91.7 \\
 & 1 & 87.8 & 94.5 & 88.2 & 94.9 & 11.1 & 18.0 & 92.0 \\
 & 2 & 89.2 & 94.3 & 88.3 & 94.9 & 11.5 & 18.3 & 92.0 \\
\multirow{-4}{*}{Same}
 & 4 & 89.2 & 94.3 & 88.5 & 94.9 & 11.7 & 18.3 & 92.1 \\
\cmidrule(l){1-9}

 & 0 & 92.3 & 97.5 & 93.6 & 93.8 & 10.1 & 18.9 & 92.5 \\
 & 1 & 95.0 & 98.6 & 94.6 & 94.0 & 11.0 & 18.7 & 92.7 \\
 & 2 & 96.3 & 98.5 & 94.7 & 94.1 & 11.5 & 19.0 & 92.8 \\
\multirow{-4}{*}{Background}
 & 4 & 96.3 & 98.5 & 94.8 & 94.1 & 11.4 & 18.9 & 92.7 \\
\cmidrule(l){1-9}

 & 0 & 85.3 & 95.5 & 89.0 & 94.4 & 12.6 & 21.4 & 90.9 \\
 & 1 & 88.0 & 96.6 & 90.0 & 94.8 & 13.9 & 21.6 & 91.4 \\
 & 2 & 89.4 & 96.6 & 90.2 & 94.7 & 14.2 & 21.8 & 91.2 \\
\multirow{-4}{*}{Illumination}
 & 4 & 89.7 & 96.7 & 90.2 & 94.7 & 14.1 & 21.8 & 91.2 \\
\cmidrule(l){1-9}

 & 0 & 87.6 & 96.1 & 92.2 & 92.4 & 14.1 & 21.7 & 88.9 \\
 & 1 & 90.0 & 97.1 & 93.2 & 92.7 & 15.2 & 22.1 & 89.1 \\
 & 2 & 91.5 & 97.0 & 93.2 & 92.7 & 15.6 & 22.1 & 89.2 \\
\multirow{-4}{*}{View}
 & 4 & 91.6 & 97.3 & 93.3 & 92.7 & 15.6 & 22.2 & 89.2 \\
\midrule

\rowcolor{oursbg}
 & 0 & 87.6 & 95.6 & 90.5 & 93.8 & 11.7 & 20.0 & 91.0 \\
\rowcolor{oursbg}
 & 1 & 90.2 & 96.7 & 91.5 & 94.1 & 12.8 & 20.1 & 91.3 \\
\rowcolor{oursbg}
 & 2 & 91.6 & 96.6 & 91.6 & 94.1 & 13.2 & 20.3 & 91.3 \\
\rowcolor{oursbg}
\multirow{-4}{*}{Mean}
 & 4 & 91.7 & 96.7 & 91.7 & 94.1 & 13.2 & 20.3 & 91.3 \\
\bottomrule
\end{tabular}
\end{minipage}
\hfill
\begin{minipage}[t]{0.49\textwidth}
\centering
\scriptsize
\setlength{\tabcolsep}{2.4pt}
\renewcommand{\arraystretch}{1.05}
\setlength{\aboverulesep}{1.2pt}
\setlength{\belowrulesep}{1.2pt}

\begin{tabular}{@{}lc ccc cccc@{}}
\toprule
\multicolumn{9}{c}{\textbf{RobustAD}~\cite{RobustAD}} \\
\midrule
 & & \multicolumn{3}{c}{Image-level}
     & \multicolumn{4}{c}{Pixel-level} \\
\cmidrule(lr){3-5}\cmidrule(l){6-9}
Category & $k$ & AUROC & AP & F1
         & AUROC & AP & F1 & AUPRO \\
\midrule

 & 0 & 90.7 & 84.5 & 81.2 & 99.1 & 16.5 & 25.2 & 96.4 \\
 & 1 & 90.8 & 84.3 & 80.5 & 99.1 & 16.4 & 24.9 & 96.3 \\
 & 2 & 91.1 & 84.5 & 81.0 & 99.1 & 16.5 & 25.0 & 96.5 \\
\multirow{-4}{*}{MetalParts}
 & 4 & 91.5 & 84.8 & 81.3 & 99.2 & 16.5 & 24.9 & 96.5 \\
\cmidrule(l){1-9}

 & 0 & 84.9 & 84.3 & 77.3 & -- & -- & -- & -- \\
 & 1 & 91.0 & 89.7 & 83.8 & -- & -- & -- & -- \\
 & 2 & 92.9 & 91.3 & 86.8 & -- & -- & -- & -- \\
\multirow{-4}{*}{PiledBags}
 & 4 & 94.8 & 93.0 & 89.8 & -- & -- & -- & -- \\
\cmidrule(l){1-9}

 & 0 & 79.9 & 72.8 & 74.6 & 97.5 & 17.3 & 26.5 & 88.7 \\
 & 1 & 82.3 & 75.6 & 75.6 & 97.8 & 18.3 & 27.8 & 89.8 \\
 & 2 & 83.6 & 78.1 & 77.1 & 97.9 & 19.0 & 28.6 & 90.6 \\
\multirow{-4}{*}{PCB}
 & 4 & 87.6 & 82.7 & 80.7 & 98.3 & 20.7 & 30.2 & 92.2 \\
\midrule

\rowcolor{oursbg}
 & 0 & 85.4 & 80.7 & 77.9 & 98.4 & 16.9 & 25.8 & 92.8 \\
\rowcolor{oursbg}
 & 1 & 88.2 & 83.3 & 80.0 & 98.5 & 17.3 & 26.2 & 93.3 \\
\rowcolor{oursbg}
 & 2 & 89.3 & 84.6 & 81.6 & 98.6 & 17.7 & 26.6 & 93.8 \\
\rowcolor{oursbg}
\multirow{-4}{*}{Mean}
 & 4 & 91.3 & 86.7 & 83.8 & 98.8 & 18.4 & 27.4 & 94.5 \\
\bottomrule
\end{tabular}
\end{minipage}

\end{table*}

\subsection{Per-domain and Detailed Results}
\Cref{tab:backbone_detailed} reports the full results of \cref{tab:backbone}.
\Cref{tab:percat} gives the per-domain results of Renormaly for \cref{tab:main}.
We also report the per-category results of Renormaly on VisA~\cite{Spot}, MVTec-AD~\cite{MVTecAD} and Real-IAD~\cite{RealIAD} (\cref{tab:conv}) in \cref{tab:percat_visa,tab:percat_mvtec,tab:percat_realiad}.

\begin{table*}[t]
\centering
\small
\caption{\textbf{Per-class performance of Renormaly on MVTec-AD~\cite{MVTecAD} Dataset for Single-Class Anomaly Detection.}}
\label{tab:percat_mvtec}
\begin{tabularx}{\textwidth}{
    @{}l*{7}{>{\centering\arraybackslash}X}@{}
}
\toprule[1.5pt]
& \multicolumn{3}{c}{Image-level}
& \multicolumn{4}{c}{Pixel-level} \\
\cmidrule(lr){2-4}\cmidrule(lr){5-8}
Category & AUROC & AP & $F_1$ & AUROC & AP & $F_1$ & AUPRO \\
\midrule
bottle       & 100.0 & 100.0 & 100.0 & 98.5 & 80.9 & 76.4 & 96.7 \\
cable        & 98.3 & 98.8 & 94.5 & 97.4 & 53.3 & 60.4 & 90.9 \\
capsule      & 97.8 & 99.4 & 96.3 & 98.4 & 43.7 & 47.2 & 97.0 \\
carpet       & 100.0 & 100.0 & 100.0 & 99.3 & 69.7 & 66.9 & 96.5 \\
grid         & 99.8 & 99.9 & 99.1 & 98.6 & 33.3 & 38.5 & 95.4 \\
hazelnut     & 100.0 & 100.0 & 100.0 & 98.7 & 57.9 & 64.2 & 94.4 \\
leather      & 100.0 & 100.0 & 100.0 & 99.0 & 43.0 & 44.5 & 98.6 \\
metal\_nut   & 100.0 & 100.0 & 100.0 & 98.3 & 79.2 & 81.6 & 89.6 \\
pill         & 98.4 & 99.6 & 96.8 & 97.7 & 70.5 & 64.8 & 97.0 \\
screw        & 97.2 & 98.8 & 95.8 & 99.2 & 31.1 & 38.7 & 97.9 \\
tile         & 99.5 & 99.7 & 98.2 & 97.6 & 63.1 & 67.1 & 89.7 \\
toothbrush   & 100.0 & 100.0 & 100.0 & 98.7 & 36.0 & 51.3 & 89.7 \\
transistor   & 99.5 & 97.9 & 94.0 & 93.8 & 59.3 & 55.9 & 83.5 \\
wood         & 98.9 & 99.7 & 96.7 & 96.9 & 60.5 & 61.0 & 92.5 \\
zipper       & 99.9 & 100.0 & 99.2 & 98.9 & 59.0 & 60.1 & 95.9 \\
\midrule
\rowcolor{oursbg} 
Mean
& 99.2 & 99.6 & 98.0
& 98.1 & 56.0 & 58.6
& 93.7 \\
\bottomrule[1.5pt]
\end{tabularx}
\end{table*}

    \begin{table*}[t]
    \centering
    \small
    \caption{\textbf{Per-class performance of Renormaly on VisA~\cite{Spot} Dataset for Single-Class Anomaly Detection.}}
    \label{tab:percat_visa}
    
    \begin{tabularx}{\textwidth}{
        @{}l*{7}{>{\centering\arraybackslash}X}@{}
    }
    \toprule[1.5pt]
    & \multicolumn{3}{c}{Image-level}
    & \multicolumn{4}{c}{Pixel-level} \\
    \cmidrule(lr){2-4}\cmidrule(lr){5-8}
    Category & AUROC & AP & $F_1$ & AUROC & AP & $F_1$ & AUPRO \\
    \midrule
    candle      & 98.7 & 99.0 & 94.6 & 99.5 & 34.0 & 44.0 & 97.1 \\
    capsules    & 99.3 & 99.5 & 98.1 & 98.4 & 43.6 & 49.5 & 96.7 \\
    cashew      & 98.2 & 99.2 & 96.3 & 98.0 & 47.6 & 50.4 & 97.1 \\
    chewinggum  & 99.2 & 99.6 & 97.3 & 99.5 & 64.1 & 64.9 & 95.6 \\
    fryum       & 99.7 & 99.9 & 99.5 & 96.6 & 39.6 & 47.0 & 94.7 \\
    macaroni1   & 98.4 & 98.2 & 94.8 & 99.8 & 26.3 & 31.9 & 97.6 \\
    macaroni2   & 95.4 & 94.9 & 89.2 & 99.7 & 14.0 & 21.4 & 98.1 \\
    pcb1        & 98.9 & 98.9 & 96.6 & 99.6 & 76.1 & 74.1 & 96.0 \\
    pcb2        & 98.9 & 99.0 & 95.7 & 98.6 & 27.4 & 37.3 & 92.8 \\
    pcb3        & 99.8 & 99.8 & 98.0 & 98.5 & 43.2 & 50.2 & 95.7 \\
    pcb4        & 99.9 & 99.9 & 98.6 & 98.5 & 39.4 & 45.0 & 90.6 \\
    pipe\_fryum & 99.9 & 99.9 & 99.0 & 98.9 & 48.1 & 56.9 & 97.8 \\
    \midrule
    \rowcolor{oursbg}
    Mean
    & 98.9 & 99.0 & 96.5
    & 98.8 & 41.9 & 47.7
    & 95.8 \\
    \bottomrule[1.5pt]
    \end{tabularx}
    \end{table*}
    
\begin{table*}[t]
\centering
\small
\caption{\textbf{Per-class performance of Renormaly on Real-IAD~\cite{RealIAD} Dataset for Single-Class Anomaly Detection.}}
\label{tab:percat_realiad}

\begin{tabularx}{\textwidth}{
    @{}l*{7}{>{\centering\arraybackslash}X}@{}
}
\toprule[1.5pt]
& \multicolumn{3}{c}{Image-level}
& \multicolumn{4}{c}{Pixel-level} \\
\cmidrule(lr){2-4}\cmidrule(lr){5-8}
Category & AUROC & AP & $F_1$ & AUROC & AP & $F_1$ & AUPRO \\
\midrule
audiojack           & 88.8 & 84.5 & 74.1 & 99.6 & 49.8 & 52.0 & 96.2 \\
bottle\_cap         & 95.1 & 93.9 & 85.0 & 99.9 & 39.1 & 42.2 & 98.9 \\
button\_battery     & 86.6 & 89.8 & 81.3 & 99.4 & 61.7 & 59.6 & 95.4 \\
end\_cap            & 87.7 & 88.7 & 84.1 & 99.2 & 25.3 & 32.5 & 97.2 \\
eraser              & 94.3 & 92.0 & 82.6 & 99.8 & 45.0 & 47.9 & 98.9 \\
fire\_hood          & 93.7 & 89.0 & 82.2 & 99.8 & 46.0 & 49.2 & 98.0 \\
mint                & 85.1 & 85.8 & 75.7 & 99.2 & 25.3 & 31.8 & 93.6 \\
mounts              & 87.1 & 73.0 & 76.0 & 99.7 & 35.3 & 38.7 & 98.4 \\
pcb                 & 92.0 & 94.9 & 86.9 & 99.7 & 50.7 & 53.6 & 97.2 \\
phone\_battery      & 93.9 & 91.8 & 83.1 & 99.7 & 45.2 & 48.3 & 98.7 \\
plastic\_nut        & 94.4 & 89.7 & 81.7 & 99.8 & 41.9 & 42.8 & 99.1 \\
plastic\_plug       & 93.5 & 90.2 & 81.0 & 99.8 & 33.2 & 39.1 & 98.8 \\
porcelain\_doll     & 91.7 & 85.7 & 77.2 & 99.7 & 35.5 & 40.5 & 98.6 \\
regulator           & 87.1 & 78.0 & 68.8 & 99.7 & 35.5 & 45.6 & 97.9 \\
rolled\_strip\_base & 99.2 & 99.3 & 96.5 & 99.9 & 45.8 & 55.6 & 99.4 \\
sim\_card\_set      & 97.6 & 97.6 & 91.9 & 99.7 & 59.9 & 57.3 & 97.9 \\
switch              & 96.7 & 96.1 & 91.7 & 98.1 & 49.9 & 58.3 & 97.3 \\
tape                & 97.5 & 95.8 & 88.6 & 99.8 & 46.3 & 51.1 & 99.1 \\
terminalblock       & 98.1 & 98.2 & 93.2 & 99.9 & 49.0 & 51.1 & 99.4 \\
toothbrush          & 85.1 & 87.7 & 78.4 & 96.3 & 27.3 & 35.1 & 86.7 \\
toy                 & 87.5 & 91.5 & 81.6 & 94.6 & 25.8 & 34.2 & 91.7 \\
toy\_brick          & 84.5 & 82.2 & 73.0 & 98.8 & 38.6 & 45.0 & 91.9 \\
transistor1         & 98.2 & 98.3 & 93.3 & 99.6 & 42.3 & 49.4 & 98.2 \\
u\_block            & 94.5 & 91.5 & 82.8 & 99.8 & 51.7 & 55.7 & 98.4 \\
usb                 & 95.8 & 94.9 & 87.8 & 99.7 & 41.5 & 46.0 & 98.9 \\
usb\_adaptor        & 88.6 & 84.1 & 76.1 & 99.7 & 27.0 & 32.9 & 97.8 \\
vcpill              & 93.5 & 92.2 & 83.6 & 99.3 & 60.2 & 64.3 & 95.9 \\
wooden\_beads       & 90.1 & 89.0 & 79.0 & 99.5 & 49.2 & 52.5 & 96.4 \\
woodstick           & 89.5 & 80.8 & 73.7 & 99.6 & 55.3 & 56.9 & 96.1 \\
zipper              & 99.2 & 99.5 & 96.0 & 99.1 & 49.7 & 55.8 & 98.2 \\
\midrule
\rowcolor{oursbg}
Mean
& 92.2 & 90.2 & 82.9
& 99.3 & 43.0 & 47.5
& 97.0 \\
\bottomrule[1.5pt]
\end{tabularx}
\end{table*}

\section{Limitations and Analysis}
\label{sec:limitation}
\Cref{fig:supp_cases} shows two successful and two failure cases at $k{=}0$. The successful cases are large defects with the largest pixel-AP gain over the control. There, the control also responds to the shift, for instance to illumination hot spots on the blade or to texture across the rotated board, and the calibration lowers these responses below the defect. The failure cases are very small defects, below 0.05\% of the image, which is smaller than a single $16{\times}16$-pixel token. The raw token residual shows that the token containing such a defect is still among the strongest responses, so the defect is located at token level. Most pixels of that token are normal, however, which bounds the pixel AP of any token-level map regardless of calibration. Two effects add to this. Normal structures such as drilled holes can respond as strongly as the defect, and the calibration can also weaken the defect response. Finer tokens or a pixel-level refinement head could address this limitation, which we leave to future work.

\end{document}